\documentclass{article} 
\usepackage{iclr2024,times}

\usepackage{hyperref}
\usepackage{url}
\usepackage{amsmath,amsfonts,bm}

\usepackage{booktabs}
\usepackage[final]{pdfpages}
\usepackage{xspace, multirow}
\usepackage{flushend}
\usepackage[font=bf]{caption}
\usepackage[ruled, vlined, linesnumbered]{algorithm2e}
\usepackage{colortbl}
\usepackage{bbold}
\usepackage{enumitem}

\usepackage{tikz}
\usepackage{pgfplots}
\usetikzlibrary{patterns}
\pgfplotsset{compat=1.16}
\usepackage[captionskip=-4pt]{subfig}
\usepackage{wrapfig}

\newcommand{\ie}{{i.e.,}\xspace}
\newcommand{\eg}{{e.g.,}\xspace}
\newcommand{\spara}[1]{\vspace{1mm}\noindent\textbf{#1.}}
\newcommand{\savespace}[1]{\ignorespaces}

\newcommand{\SGC}{SGC\xspace}
\newcommand{\SIGN}{SIGN\xspace}
\newcommand{\ChebNet}{ChebNet\xspace}
\newcommand{\ChebNetII}{ChebNetII\xspace}
\newcommand{\GPRGNN}{GPR-GNN\xspace}
\newcommand{\BernNet}{BernNet\xspace}
\newcommand{\JacobiConv}{JacobiConv\xspace}
\newcommand{\ASGC}{ASGC\xspace}
\newcommand{\OptBasisGNN}{OptBasisGNN\xspace}
\newcommand{\newbasis}{UniBasis\xspace}
\newcommand{\ours}{UniFilter\xspace}

\newcommand{\GCN}{GCN\xspace}
\newcommand{\GCNII}{GCNII\xspace}
\newcommand{\GAT}{GAT\xspace}
\newcommand{\MixHop}{MixHop\xspace}
\newcommand{\EvenNet}{EvenNet\xspace}
\newcommand{\HGCN}{H$_2$GCN\xspace}
\newcommand{\LINKX}{LINKX\xspace}
\newcommand{\ACM}{ACM-GCN\xspace}
\newcommand{\GloGNN}{GloGNN++\xspace}
\newcommand{\WRGAT}{WRGAT\xspace}
\newcommand{\Specformer}{Specformer\xspace}
\newcommand{\OrderedGNN}{Orderred GNN\xspace}

\newcommand{\HomFilter}{HomFilter\xspace}
\newcommand{\HetFilter}{HetFilter\xspace}
\newcommand{\OrtFilter}{OrtFilter\xspace}

\newcommand{\A}{\mathbf{A}\xspace}

\newcommand{\D}{\mathbf{D}\xspace}
\newcommand{\G}{\mathbf{G}\xspace}

\newcommand{\I}{\mathbf{I}\xspace}
\renewcommand{\P}{\mathbf{P}\xspace}
\newcommand{\U}{\mathbf{U}\xspace}

\newcommand{\X}{\mathbf{X}\xspace}
\newcommand{\Y}{\mathbf{Y}\xspace}

\newcommand{\bLambda}{\mathbf{\Lambda}\xspace}

\renewcommand{\L}{\mathbf{L}\xspace}

\newcommand{\TD}{\tilde{\D}\xspace}
\newcommand{\TA}{\tilde{\A}\xspace}

\newcommand{\Tlambda}{\tilde{\lambda}\xspace}

\newcommand{\h}{\mathbf{h}\xspace}

\newcommand{\s}{\mathbf{s}\xspace}

\newcommand{\w}{\mathbf{w}\xspace}
\newcommand{\x}{\mathbf{x}\xspace}
\newcommand{\y}{\mathbf{y}\xspace}
\renewcommand{\v}{\mathbf{v}\xspace}
\renewcommand{\u}{\mathbf{u}\xspace}
\newcommand{\z}{\mathbf{z}\xspace}

\newcommand{\C}{\mathcal{C}\xspace}
\newcommand{\R}{\mathbb{R}\xspace}
\newcommand{\N}{\mathcal{N}\xspace}
\newcommand{\F}{\mathcal{F}\xspace}

\newcommand{\V}{\mathcal{V}\xspace}
\newcommand{\E}{\mathcal{E}\xspace}

\newcommand{\g}{\ensuremath{\mathrm{g}}}

\newtheorem{definition}{Definition}
\newtheorem{theorem}{Theorem}
\newtheorem{lemma}{Lemma}[section]

\newtheorem{proof}{Proof}

\newcommand{\eat}[1]{}
\newcommand{\revise}[1]{{#1}}
\newcommand{\del}[1]{}

\title{An Effective Universal Polynomial Basis for Spectral Graph Neural Networks}

\author{Keke Huang \\
National University of Singapore \\
\texttt{kkhuang@nus.edu.sg} 
\And
Pietro Li\`{o} \\
University of Cambridge\\
\texttt{pl219@cam.ac.uk}
}

\iclrfinalcopy 
\begin{document}

\maketitle

\begin{abstract}
\eat{They have been extensively explored in the spectral perspective, referred to as {\em graph filters}.}
Spectral Graph Neural Networks (GNNs), also referred to as {\em graph filters} have gained increasing prevalence for heterophily graphs. Optimal graph filters rely on Laplacian eigendecomposition for Fourier transform. In an attempt to avert the prohibitive computations, numerous polynomial filters by leveraging distinct polynomials have been proposed to approximate the desired graph filters. However, polynomials in the majority of polynomial filters are {\em predefined} and remain {\em fixed} across all graphs, failing to accommodate the diverse heterophily degrees across different graphs. To tackle this issue, we first investigate the correlation between polynomial bases of desired graph filters and the degrees of graph heterophily via a thorough theoretical analysis. Afterward, we develop an adaptive heterophily basis by incorporating graph heterophily degrees. Subsequently, we integrate this heterophily basis with the homophily basis, creating a universal polynomial basis {\em \newbasis}. In consequence, we devise a general polynomial filter {\em \ours}. Comprehensive experiments on both real-world and synthetic datasets with varying heterophily degrees significantly support the superiority of \ours, demonstrating the effectiveness and generality of \newbasis, as well as its promising capability as a new method for graph analysis. 
\end{abstract}

\begin{sloppy}
\section{Introduction}\label{sec:intro}

Spectral Graph Neural Networks (GNNs)~\citep{KipfW17}, known as {\em graph filters} have been extensively investigated in recent years due to their superior performance in handling heterophily graphs. Optimal graph filters conduct Laplacian eigendecomposition for Fourier transform. To bypass the computation complexity, existing graph filters leverage various polynomials to approximate the desired filters for graphs with varying heterophily degrees. For example, \ChebNet~\citep{DefferrardBV16} employs truncated Chebyshev polynomials~\citep{mason2002chebyshev, hammond2011wavelets} and accomplishes localized spectral filtering. \BernNet~\citep{he2021bernnet} utilizes Bernstein polynomials~\citep{Farouki12} to acquire better controllability and interpretability. Later, \citet{WangZ22} propose \JacobiConv by exploiting Jacobi polynomial bases~\citep{askey1974positive} with improved generality. Recently, the state-of-the-art (SOTA) graph filter \OptBasisGNN~\citep{GuoW23} orthogonalizes the polynomial basis to reach the maximum convergence speed.

However, existing polynomial filters overlook the diverse heterophily degrees of underlying graphs when implementing polynomial bases. This oversight of the graph homophily characteristic in the development of polynomial bases leads to suboptimal performance on real-world graphs, as demonstrated in our experiments (Sections~\ref{sec:classresults} and~\ref{sec:ablation}). This scenario arises from the lack of consideration of the homophily property in the construction of the orthonormal basis, rendering it inferior to strong homophily graphs. As we prove in Theorem~\ref{thm:frequencyratio}, frequencies of signals filtered by optimal graph filters are proportional to the heterophily degrees. This suggests that ideal polynomial bases are obligated to provide adaptability to the diverse heterophily degrees. 

Ergo, a natural question to ask is: {\bf how can we design a universal polynomial basis that encapsulates the graph heterophily degrees?} Inspired, we first establish the relation between the heterophily degree and the frequency of optimal filtered signals (Theorem~\ref{thm:frequencyratio}). Subsequently, we explore how the distribution of polynomial bases in Euclidean space affects the basis spectrum (Theorem~\ref{thm:pivot}). Based on those insightful findings, we design an adaptive heterophily basis by incorporating heterophily degrees of graphs. Eventually, we integrate the heterophily basis and the homophily basis into a universal basis denoted as {\em\newbasis}. Upon \newbasis, we devise a general polynomial filter called {\em \ours}. For a comprehensive evaluation, we compare \ours with \revise{$20$} baselines on $6$ real-world datasets and synthetic datasets with a range of heterophily degrees. The notably superior performance of \ours strongly confirms the effectiveness and generality of \newbasis, especially on heterophily graphs. Meanwhile, we demonstrate the spectrum distribution of trained \newbasis on each tested dataset (section~\ref{sec:spectrum}). The experimental results explicitly support the promising capability of \newbasis as a new method for graph analysis with enriched interpretability.

In a nutshell, our contribution can be summarized as: 1) We reveal that underlying polynomials of desired polynomial filters are meant to keep aligned with degrees of graph heterophily; 2) We design a universal polynomial basis \newbasis by incorporating graph heterophily degrees and devise a general graph filter \ours; 3) We evaluate \ours on both real-world and synthetic datasets against $18$ baselines. The remarkable performance of \ours strongly confirms the effectiveness and generality of \newbasis, as well as its promising capability as a new method for graph analysis.      

\section{Preliminaries}\label{sec:preliminary}

\subsection{Notations and Definitions} 

We represent matrices, vectors, and sets with bold uppercase letters (\eg $\A$), bold lowercase letters (\eg $\x$), and calligraphic fonts (\eg $\N$), respectively. The $i$-th row (resp.\ column) of matrix $\A$ is represented by $\A[i,\cdot]$ (resp.\ $\A[\cdot, i]$). We denote $[n]=\{1,2,\cdots, n\}$.  

Let $\G=(\V, \E)$ be an undirected and connected graph with node set $|\V|=n$ and edge set $|\E|=m$. Let $\X \in \R^{n\times d}$ be the $d$-dimension feature matrix. For ease of exposition, we employ node notation $u\in \V$ to denote its index, \ie $\X_u=\X[u,\cdot]$. Let $\Y\in \mathbb{N}^{n\times |\C|}$ be the one-hot label matrix, \ie $\Y[u,i]=1$ if node $u$ belongs to class $\C_i$ for $i\in \{1,2,\cdots, |\C|\}$, where $\C$ is the set of node labels. The set of direct (one-hop) neighbors of node $u\in \V$ is denoted as $\N_u$ with degree $d_u=|\N_u|$. The adjacency matrix of $\G$ is denoted as $\A \in \R^{n\times n}$ that $\A[u,v]=1$ if edge $\langle u, v \rangle \in \E$; otherwise $\A[u,v]=0$. $\D \in \R^{n\times n}$ is the diagonal degree matrix of $\G$ with $\D[u,u]=d_u$. Let $\L$ be the normalized Laplacian matrix of graph $\G$ defined as $\L=\I-\D^{-\frac{1}{2}}\A\D^{-\frac{1}{2}}$ where $\I$ is the identity matrix and $\hat{\L}$ be the normalized Laplacian matrix of $\G$ with self-loops as $\hat{\L}=\I-\TD^{-\frac{1}{2}}\TA\TD^{-\frac{1}{2}}$ where $\TD=\D+\I$ and $\TA=\A+\I$.

\subsection{Spectral graph filters} 

In general, the eigendecomposition of the Laplacian matrix is denoted as $\L=\U \bLambda \U^\top$, where $\U$ is the matrix of eigenvectors and $\bLambda=\mathrm{diag}[\lambda_1, \cdots, \lambda_n]$ is the diagonal matrix of eigenvalues. Eigenvalues $\lambda_i$ for $i \in [n]$ mark the {\em frequency} and the eigenvalue set $\{\lambda_1, \cdots, \lambda_n\}$ is the {\em graph spectrum}. Without loss of generality, we assume $0=\lambda_1 \le \lambda_2 \le \cdots \le \lambda_n \le 2$. When applying a spectral graph filter on graph signal $\x \in \R^n$, the process involves the following steps. First, the graph Fourier operator $\F(\x)=\U^\top \x$ projects the graph signal $\x$ into the spectral domain. Subsequently,  a spectral filtering function $\g_\w(\cdot)$ parameterized by $\w\in \R^n$ is employed on the derived spectrum. Eventually, the filtered signal is transformed back via the inverse graph Fourier transform operator $\F^{-1}(\x)=\U \x$. The process is formally expressed as 
\begin{align}
&\F^{-1}(\F(\g_\w) \odot  \F(\x))=\U\g_\w(\bLambda)\U^\top \x =\U\ \mathrm{diag}(\g_\w(\lambda_1), \cdots, \g_\w(\lambda_n))\U^\top \x, \label{eqn:Fourier} 
\end{align}
where $\odot$ is the Hadamard product. 

In particular, spectral graph filters enhance signals in specific frequency ranges and suppress the signals in the rest parts according to objective functions. For node classification, homophily graphs are prone to contain low-frequency signals whilst heterophily graphs likely own high-frequency signals. In order to quantify the heterophily degrees of graphs, numerous homophily metrics have been introduced, \eg\ {\em edge homophily}~\citep{ZhuYZHAK20}, {\em node homophily}~\citep{PeiWCLY20}, {\em class homophily}~\citep{lim2021large,luan2021heterophily}, and a recent {\em adjusted homophily}~\citep{platonov2022}. By following the literature of spectral graph filters~\citep{ZhuYZHAK20, LeiWLDW22}, we adopt edge homophily in this work, explained as follows.

\begin{definition}[Homophily Ratio $h$]~\label{def:homo}
Given a graph $\G=(\V,\E)$ and its label matrix $\Y$, the homophily ration $h$ of $\G$ is the fraction of edges with two end nodes from the same class, \ie $h=\frac{|\{\langle u,v \rangle \in \E \colon \y_u=\y_v\} |}{|\E|}$.
\end{definition}

Besides the homophily metrics for {\em categorical} node labels, the similarity of {\em numerical} node signals can also be measured via {\em Dirichlet Enenrgy}~\citep{ZhouHZCLCH21, KarhadkarBM23}. Specifically, we customize the metric to node signals $\x\in \R^n$ and propose {\em spectral signal frequency} as follows. 

\eat{As indicated, the homophily ratio is to measure the consistency of label distribution on graphs. This metric is categorically defined for label signals, which, however, is not feasible for numerical feature signals. To generalize the concept of homophily ratio and quantify the {\em consistency} of feature signals, we propose the concept of {\em spectral signal frequency} as follows.}

\begin{definition}[Spectral Signal Frequency $f$]
Consider a graph $\G=(\V,\E)$ with $n$ nodes and Laplacian matrix $\L$. Given a normalized feature signal $\x\in \R^n$, the spectral signal frequency $f(\x)$ on $\G$ is defined as $f(\x)=\tfrac{\x^\top\L\x}{2}$.
\end{definition} 
By nature of Dirichlet energy, spectral signal frequency $f(x)$ quantifies the discrepancy of signal $\x$ on graph $\G$. For $f(x)$, it holds that
\begin{lemma}\label{lem:frequencybound}
For any normalized feature signal $\x \in \R^{n}$ on graph $\G$, the spectral signal frequency $f(\x)\in [0,1]$ holds.  
\end{lemma}

\section{Revisiting Polynomial graph filters}\label{sec:revisting}

Optimal graph filters require eigendecomposition on the Laplacian matrix at the cost of $O(n^3)$. To bypass the high computation overhead, a plethora of polynomial graph filters~\citep{DefferrardBV16, ChienP0M21, he2021bernnet, WangZ22, HeWW22, GuoW23} have been proposed to approximate optimal graph filters by leveraging distinct polynomials. Table~\ref{tbl:pgf} summarizes several such polynomial graph filters, including adopted polynomials, graph filter functions, and propagation matrices if applicable.

\begin{table}[!t]
\caption{Polynomial Graph Filters} \label{tbl:pgf}\vspace{-2mm}
\small
\vspace{-1mm}
\renewcommand{\arraystretch}{1.2}
\begin{tabular}{@{}l|l|l|l}
\toprule
{} & Poly. Basis & Graph Filter $\g_\w(\lambda)$ & Prop. Matrix $\P$  \\ \midrule
\ChebNet~\citep{DefferrardBV16} & Chebyshev & $\sum^K_{k=0}\w_k T_k(\hat{\lambda})$ & $2\L/\lambda_{max}-\I$ \\ 
\GPRGNN~\citep{ChienP0M21}  & Monomial & $\sum^K_{k=0}\w_k(1-\Tlambda)^k$ & $\I-\hat{\L}$ \\
\BernNet~\citep{he2021bernnet} & Bernstein & $\sum^K_{k=0}\frac{\w_k}{2^K}\binom{K}{k}(2-\lambda)^{K-k}\lambda^k$ & $\I-\frac{\L}{2}$ \\ 
\JacobiConv~\citep{WangZ22} & Jacobi & $\sum^K_{k=0}\w_k\P^{a,b}_k(1-\lambda)$ & $\I-\L$ \\ 
\OptBasisGNN~\citep{GuoW23} & Orthonormal & --- & $\I-\L$ \\  \bottomrule
\end{tabular}
\vspace{-1mm}
\end{table}

By identifying the appropriate matrix $\P$, those polynomial filters applied on graph signal $\x\in \R^n$ can be equally expressed as 
\begin{equation}\label{eqn:gsf}
\z=\textstyle\sum^K_{k=0}\w_k \P^k\cdot \x,   
\end{equation}
where $K$ is the length of polynomial basis, $\w\in \R^{K+1}$ is the learnable weight vector, and $\z\in \R^n$ is the final representation. For example, \citet{he2021bernnet} utilize Bernstein polynomial and propose polynomial filter \BernNet as $\z=\textstyle\sum^K_{k=0}\frac{\w^\prime_k}{2^K}\binom{K}{k}(2\I-\L)^{K-k}\L^k \x$. By setting $\P=\I-\frac{\L}{2}$ as the propagation matrix and rearranging the expression, an equivalent formulation is expressed as $\z=\textstyle\sum^K_{k=0}\w_k \left(\I-\frac{\L}{2}\right)^k \x$ where $\w_k=\sum^k_{i=0} \w^\prime_{k-i}\binom{K}{K-i}\binom{K-i}{k-i}(-1)^{k-i}$ is the learnable parameter. 

In particular, vectors $\P^k\x$ in Equation~\eqref{eqn:gsf} for $k\in \{0,1,\cdots, K\}$ collectively constitute a signal basis $\{\P^0\x, \P^1\x, \cdots, \P^K\x\}$. Spectral graph filters attempt to learn a weighted combination of signal bases, aiming to systematically produce node representations for nodes from graphs with varying heterophily degrees for label prediction. From the spectral perspective, spectral filters essentially execute filtering operations on the spectrum $\{f(\P^0\x), f(\P^1\x), \cdots, f(\P^K\x)\}$ in order to approximate the frequencies of label signals $\Y$. Meanwhile, label signal frequencies are closely correlated with the homophily ratio $h$. To formally depict the correlation between the filtered signal $\textstyle\sum^K_{k=0}\w_k\P^k\x$ and homophily ratio $h$, we establish a theorem as follows.

\begin{theorem}\label{thm:frequencyratio}
Given a connected graph $\G=(\V,\E)$ with homophily ratio $h$, consider an optimal polynomial filter $\mathrm{F(\w)}=\textstyle\sum^K_{k=0}\w_k \P^k$ with propagation matrix $\P$ and weights $\w \in \R^{K+1}$ toward $\G$ for node classification. Given a feature signal $\x\in \R^n$, the spectral frequency $f(\textstyle\sum^K_{k=0}\w_k\P^k\x)$ is proportional to $1\!-\!h$. 
\end{theorem}

Theorem~\ref{thm:frequencyratio} uncovers the critical role of graph homophily ratios when generating desired node representations. Intuitively, ideal signal bases are obligated to consider different heterophily degrees for various graphs. However, the majority of existing polynomial filters exploit predefined polynomials, ignoring the corresponding homophily ratios.

\section{Universal Polynomial Basis for graph filters}\label{sec:unibasis}

\begin{wrapfigure}{r}{0.6\textwidth}\vspace{-10mm}
    \centering
    \captionsetup[subfigure]{margin={0.2cm,0.1cm}}
    \subfloat[{Trend of homophily basis}]{\includegraphics[height=0.28\textwidth]{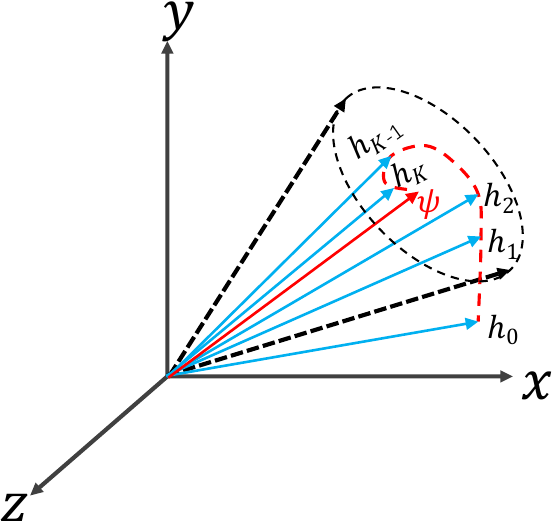}\label{subfig:homo}}\hspace{0mm}
    \subfloat[{Construction of heterophily basis}]{\includegraphics[height=0.28\textwidth]{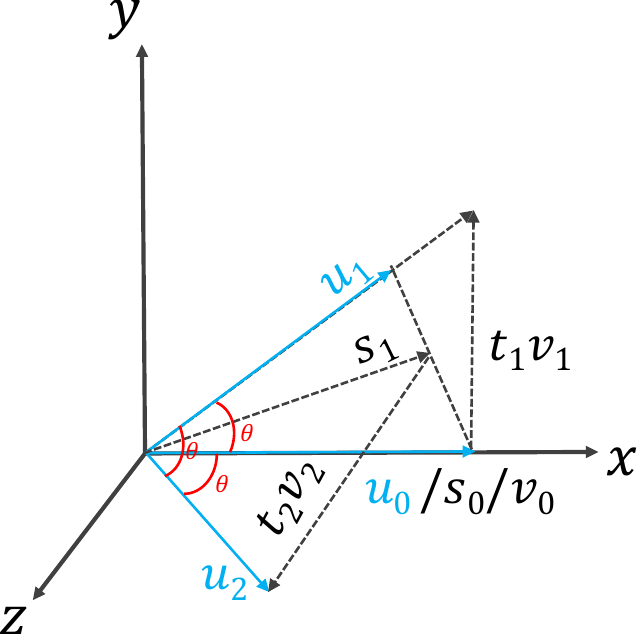}\label{subfig:hetero}}
    \caption{Illustration of homophily and heterophily bases}\label{fig:homoheterbasis}\vspace{-5mm}
\end{wrapfigure}

\subsection{Theoretical Analysis of Homophily Basis}\label{sec:homobasis}

Conventional GNN models~\citep{KipfW17, HamiltonYL17, KlicperaBG19} employ homophily as a strong inductive bias~\citep{lim2021large}. To aggregate information within $K$ hops, graph signal $\x$ are propagated to $K$-hop neighbors via propagation matrix $\P=\I-\L$, yielding {\em homophily basis} $\{\x, \P\x,\cdots,\P^K\x\}$. To elucidate how the homophily basis accommodates homophily graphs, we establish the following Theorem.
\begin{theorem}\label{thm:homo}
Given a propagation matrix $\P$ and graph signal $\x$, consider an infinite homophily basis $\{\x, \P\x, \cdots, \P^k\x, \P^{k+1}\x, \cdots\}$. There exists an integer $\eta \in \mathbb{N}$ such that when the exponent $k\ge \eta$ and increases, the angle $\arccos\left(\tfrac{\P^k\x \cdot \P^{k+1}\x}{\|\P^k\x\| \|\P^{k+1}\x\|}\right)$ is progressively smaller and asymptotically approaches $0$.
\end{theorem}

The homophily basis exhibits {\em growing similarity} and {\em asymptotic convergence} for the purpose of capturing homophily signals, thus resulting in the {\em over-smoothing issue}. For better visualization, Figure~\ref{subfig:homo} simply illustrates the homophily basis $\{\h_0, \h_1, \h_2, \cdots, \h_{K-1}, \h_K, \cdots\}$ gradually converges to $\psi$ in $3$-dimension Euclidean space.

\subsection{Adaptive Heterophily Basis}\label{sec:heterobasis}

\eat{Spectral graph filters aim to systematically produce optimal node representations for graphs with varying heterophily degrees from a signal basis constructed from graph signals and structures. As proved in Theorem~\ref{thm:frequencyratio}, the frequency of the optimal node presentation (signal) is aligned with the homophily ratios of graphs. The above two observations suggest that}

As discussed aforementioned, desired signal bases are expected to conform to homophily ratios. A natural question is: {\em how can we apply homophily ratios in a sensible manner when designing signal bases without involving graph signals or structures?} To answer this question, we initially explore the correlation between the basis distribution in Euclidean space and the basis frequency on regular graphs.

\begin{theorem}\label{thm:pivot}
Consider a regular graph $\G$, a random basis signal $\x \in \R^n$, and a normalized all-ones vector $\phi\in \R^n$ \revise{with frequency $f(\phi)=0$}. Suppose $\theta:=\mathrm{arccos}( \phi \cdot \x )$ denotes the angle formed by $\x$ and $\phi$. It holds that the expectation of spectral signal frequency $\mathbb{E}_{\G\sim \mathcal{G}}[f(\x)]$ over the randomness of $\G$ is monotonically increasing with $\theta$ for $\theta \in [0,\tfrac{\pi}{2})$.
\end{theorem}

\revise{Theorem~\ref{thm:pivot} reveals the correlation between the expected frequency of the signal basis and its relative position to the $0$-frequency vector $\phi$ on regular graphs. This fact implicitly suggests that we may take the angles (relative position) between two basis vectors into consideration when aiming to achieve the desired basis spectrum on general graphs. Meanwhile, Theorem~\ref{thm:homo} discloses the growing similarity and asymptotic convergence phenomenon within the homophily basis. To mitigate this over-smoothing issue, we can intuitively enforce all pairs of basis vectors to form an appropriate angle of $\theta \in [0,\tfrac{\pi}{2}]$. Pertaining to this, Theorem~\ref{thm:frequencyratio} proves the spectral frequency of ideal signals proportional to $1-h$, aligning with the homophily ratios of the underlying graphs. By leveraging the monotonicity property proved in Theorem~\ref{thm:pivot}, we empirically set the $\theta:=\frac{\pi}{2}(1-h)$.} Consequently, a signal basis capable of capturing the heterophily degrees of graphs is derived, formally denoted as {\em heterophily basis}.

Consider to construct a heterophily basis with a length of $K\!+\!1$. The procedure of computing heterophily basis is outlined in Algorithm~\ref{alg:basis} and illustrated in Figure~\ref{subfig:hetero}. To start with, we normalize the input signal $\x$ as the initial signal $\u_0$ and set $\theta:=\tfrac{(1-h)\pi}{2}$. In order to manipulate the formed angles between signal vectors, we forge an orthonormal basis, denoted as $\{\v_0, \v_1, \cdots, \v_K\}$ where $\v_0$ is initialized as $\u_0$. In particular, at the $k$-th iteration for $k\in[1, K]$, we set $\v_k= \P \v_{k-1}$ where $\P= \I-\L$ is the propagation matrix. Subsequently, $\v_k$ is calculated as $\v_k:= \v_k-  (\v^\top_k \v_{k-1})\v_{k-1} -  (\v^\top_k \v_{k-2}) \v_{k-2}$ as per the {\em three-term recurrence} Theorem~\citep{gautschi2004orthogonal,liesen2013krylov,GuoW23}. Meanwhile, signal vector $\u_k$ is set as $\u_k:=\tfrac{\s_{k-1}}{k}$ where $\s_{k-1}:=\sum^{k-1}_{i=0}\u_i$. Subsequently, $\u_k$ is updated as $\u_k := \tfrac{\u_k+t_k\v_k}{\|\u_k+t_k\v_k\|}$ where $t_k$ is 
\begin{equation}\label{eqn:factor}
t_k=\sqrt{\left(\tfrac{\s_{k-1}^\top\u_{k-1}}{k\mathrm{cos}(\theta)}\right)^2-\tfrac{(k-1)\mathrm{cos(\theta)}+1}{k}}.    
\end{equation}
As a result, the final vector set $\{\u_0,\u_1,\cdots,\u_K\}$ is returned as the heterophily basis. The desired property of the heterophily basis is proved in the following Theorem. Detailed proofs are presented in Appendix~\ref{app:proofs}.

\begin{algorithm}
\caption{Heterophily Basis}
\label{alg:basis}
\KwIn{Graph $\G$, propagation matrix $\P$, input feature signal $\x$, hop $K$, \revise{estimated homophily ratio $\hat{h}$}}
\KwOut{Heterophily basis $\{\u_0,\u_1,\cdots,\u_K\}$}
$\u_0\gets \tfrac{\x}{\|\x\|}$, $\v_0 \gets \u_0$, $\v_{-1} \gets \mathbf{0}$, $\s_0\gets \u_0$, \revise{$\theta \gets \tfrac{(1-\hat{h})\pi}{2}$}\;
\For{$k \gets 1$ \KwTo $K$} 
{
    $\v_k\gets \P\v_{k-1}$\;
    $\v_k \gets \v_k-  (\v^\top_k \v_{k-1})\v_{k-1} -  (\v^\top_k \v_{k-2}) \v_{k-2}$\;
    $\v_k \gets \tfrac{\v_k}{\|\v_k\|}$, $\u_k\gets \textstyle\tfrac{\s_{k-1}}{k}$\;
    $t_k$ is calculated as in Equation~\eqref{eqn:factor}\;
    $\u_k \gets \tfrac{\u_k+t_k\v_k}{\|\u_k+t_k\v_k\|}$, $\s_k \gets \s_{k-1}+\u_k$\;
}
\Return $\{\u_0,\u_1,\cdots,\u_K\}$\;
\end{algorithm}

\begin{theorem}\label{thm:heteroproperty}
Consider a heterophily basis $\{\u_0,\u_1, \cdots \u_K\}$ constructed from Algorithm~\ref{alg:basis} for graphs with homophily ratio $h$. It holds that
$\u_i \cdot \u_j =\begin{cases}
   \mathrm{cos}(\tfrac{(1-h)\pi}{2}) & \textrm{if } i\neq j \\
   1 & \textrm{if } i = j
\end{cases}$
for $\forall i,j\in \{0,1,\cdots,K\}$.
\end{theorem}

\revise{\spara{Homophily ratio estimation} The exact homophily ratio $h$ relies on the label set of the entire graph and thus is normally unavailable. To address this issue, we estimate $h$ through labels of training data, denoted as $\hat{h}$. Appendix~\ref{app:exp} presents the experimental results of the homophily ratio estimation, which signifies that a qualified homophily ratio can be effectively estimated via training data.}

\spara{Time complexity} In the $k$-th iteration, it takes $O(m+n)$ to calculate the orthonormal basis and $O(n)$ to update $\u_k$. Therefore, the total time complexity of Algorithm~\ref{alg:basis} is $O(K(m+n))$, \ie linear to propagation hops and input graph sizes.

\subsection{Universal polynomial basis and graph filter}\label{sec:ourmodel}

The heterophily basis employs fixed angle \revise{$\theta:=\tfrac{(1-\hat{h})\pi}{2}$} associated with heterophily degrees, effectively encapsulating the heterophily of graphs. However, it can be restrictive by nature when handling strong homophily graphs with homophily ratios $h$ close to $1$. To tackle graphs ranging from strong homophily to strong heterophily, we intuitively introduce a hyperparameter $\tau\in[0,1]$ and merge the homophily basis and heterophily basis into a universal polynomial $\tau\P^k \x+(1-\tau)\u_k$, referred to as {\em \newbasis}. As a consequence, a general polynomial filter {\em \ours} is proposed as 
\begin{equation}\label{eqn:generalbasis}
\z=\textstyle\sum^K_{k=0}\w_k (\tau\P^k \x+(1-\tau)\u_k) 
\end{equation}
with learnable weight vector $\w \in \R^{K+1}$.


\section{Experiments}\label{sec:exp}

\spara{Datasets} We evaluate the performance of \ours on $6$ real-world datasets with varied homophily ratios. Specifically, the three citation networks~\citep{sen2008collective}, \ie Cora, Citeseer, and Pubmed, are homophily graphs with homophily ratios $0.81$, $0.73$, and $0.80$ respectively; the two Wikipedia graphs, \ie Chameleon and Squirrel and the Actor co-occurrence graph from WebKB3~\citep{PeiWCLY20} are heterophily graphs with homophily ratios $0.22$, $0.23$, and $0.22$ respectively. Dataset details are presented in Table~\ref{tbl:dataset} in Appendix~\ref{app:settings}.

\spara{Baselines} We compare \ours with \revise{$20$} baselines in two categories, \ie\ {\em polynomial filters} and {\em model-optimized methods}. Specifically, polynomial filters employ various polynomials to approximate the optimal graph filters, including monomial \SGC~\citep{WuSZFYW19}, \SIGN~\citep{frasca2020sign}, \revise{\ASGC~\cite{ChanpuriyaM22}}, \GPRGNN~\citep{ChienP0M21}, and \EvenNet~\citep{LeiWLDW22}, Chebyshev polynomial \ChebNet~\citep{DefferrardBV16} and its improved version \ChebNetII~\citep{HeWW22}, Bernstein polynomial \BernNet~\citep{he2021bernnet}, Jacobi polynomial \JacobiConv~\citep{WangZ22}, the orthogonal polynomial \OptBasisGNN~\citep{GuoW23} and learnable basis \Specformer~\citep{BoSWL23}. In contrast, model-optimized methods optimize the architecture for improved node representations, including \GCN~\citep{KipfW17}, \GCNII~\citep{ChenWHDL20}, \GAT~\citep{VelickovicCCRLB18}, \MixHop~\citep{AbuElHaijaPKA19}, \HGCN~\citep{ZhuYZHAK20}, \LINKX~\citep{lim2021large}, \revise{\WRGAT~\citep{SureshBNLM21}}, \ACM~\citep{LuanHLZZZCP22}, and \GloGNN~\citep{LiZCSLLQ22}. 

\spara{Experiment Settings} There are two common data split settings $60\%/20\%/20\%$ and $48\%/32\%/20\%$ for train/validation/test in the literature. Specifically, the polynomial filters are mostly tested in the previous setting~\citep{he2021bernnet,WangZ22,GuoW23,BoSWL23} while the model-optimized methods are normally evaluated in the latter\footnote{Please note that those model-optimized methods reuse the public data splits from~\citet{PeiWCLY20} which are actually in the splits of $48\%/32\%/20\%$ in the implementation.}~\citep{ZhuYZHAK20,LiZCSLLQ22,SongZWL23}.

\subsection{Node Classification Performance}\label{sec:classresults}

\begin{table}[!t]
 \caption{Accuracy (\%) compared with polynomial filters}\label{tbl:polyfilter}\vspace{-1mm}
 \small
\resizebox{\textwidth}{!}{%
\begin{tabular}{@{}c|c|c|c|c|c|c@{}}
\toprule
\multicolumn{1}{c}{\bf Methods} & \multicolumn{1}{c}{\bf Cora} & \multicolumn{1}{c}{\bf Citeseer} & \multicolumn{1}{c} {\bf Pubmed} &\multicolumn{1}{c}{\bf Actor} & \multicolumn{1}{c} {\bf Chameleon}& \multicolumn{1}{c}{\bf Squirrel} \\ \midrule
\SGC             &	   86.83 $\pm$ 1.28 	&	   79.65 $\pm$ 1.02  	&	   87.14 $\pm$ 0.90    	 &	34.46 $\pm$ 0.67    &	   44.81 $\pm$ 1.20    &	 25.75 $\pm$ 1.07    \\
 \SIGN                   &	87.70 $\pm$ 0.69  	&	80.14 $\pm$ 0.87   	&	 89.09 $\pm$ 0.43      & 		41.22 $\pm$ 0.96      	&	60.92 $\pm$ 1.45         	&	45.59 $\pm$ 1.40    	\\
 \revise{\ASGC} & \revise{85.35 $\pm$ 0.98} & \revise{76.52 $\pm$ 0.36} & \revise{84.17 $\pm$ 0.24} &  \revise{33.41 $\pm$ 0.80} & \revise{71.38 $\pm$ 1.06} & \revise{57.91 $\pm$ 0.89} \\
 \GPRGNN               	&	   88.54 $\pm$ 0.67  	&	   80.13 $\pm$ 0.84  	&	   88.46 $\pm$ 0.31     &	   39.91 $\pm$ 0.62  	&	   67.49 $\pm$ 1.38     	&	   50.43 $\pm$ 1.89    \\
\EvenNet               &  87.77 $\pm$ 0.67   &	78.51 $\pm$ 0.63     	&\underline{90.87 $\pm$ 0.34} &  	40.36 $\pm$ 0.65       	&	67.02 $\pm$ 1.77         	&	52.71 $\pm$ 0.85  	\\
 \ChebNet              	&	   87.32 $\pm$ 0.92  	&	   79.33 $\pm$ 0.57  	&	   87.82 $\pm$ 0.24  	&	   37.42 $\pm$ 0.58   	&	   59.51 $\pm$ 1.25     	&	   40.81 $\pm$ 0.42     \\
 \ChebNetII             & 88.71  $\pm$ 0.93	    &	80.53  $\pm$ 0.79     	&	88.93  $\pm$ 0.29    &  	41.75 $\pm$ 1.07 &	71.37  $\pm$ 1.01        	&	57.72  $\pm$ 0.59    	\\
 \BernNet              	&	   88.51 $\pm$ 0.92  	&	   80.08 $\pm$ 0.75  	&	   88.51 $\pm$ 0.39    &	   41.71 $\pm$ 1.12 &	   68.53 $\pm$ 1.68     	&	   51.39 $\pm$ 0.92    	\\
 \JacobiConv           	&	   \underline{88.98 $\pm$ 0.72}  	&	   80.78 $\pm$ 0.79    	&	   89.62 $\pm$ 0.41  	&	   41.17 $\pm$ 0.64   	&	   74.20 $\pm$ 1.03     &	   57.38 $\pm$ 1.25   \\
 \OptBasisGNN           & 87.00 $\pm$ 1.55	    &	80.58  $\pm$ 0.82     	&	90.30  $\pm$ 0.19 &	\bf{42.39 $\pm$ 0.52} &	74.26 $\pm$ 0.74      	&	63.62  $\pm$ 0.76   	\\
 \Specformer    &	88.57 $\pm$ 1.01   & \underline{81.49 $\pm$ 0.94} & 87.73 $\pm$ 0.58    &  \underline{41.93 $\pm$ 1.04} & \underline{74.72 $\pm$ 1.29}    &  \underline{64.64 $\pm$ 0.81}	\\
 \ours      &	\revise{\bf{89.49 $\pm$ 1.35}}   &  \revise{\bf{81.39 $\pm$ 1.32}} & \revise{\bf{91.44 $\pm$ 0.50}}  & \revise{40.84 $\pm$ 1.21} &  \revise{\bf{75.75 $\pm$ 1.65}}  & \revise{\bf{67.40 $\pm$ 1.25}}    	\\ \bottomrule 
\end{tabular}}
\end{table}
\vspace{-3mm} 
\begin{table}[!t]
 \caption{Accuracy (\%) compared with model-optimized methods}\label{tbl:modelopt}\vspace{-1mm}
 \small
\resizebox{\textwidth}{!}{%
\begin{tabular}{@{}c|c|c|c|c|c|c@{}}
\toprule
\multicolumn{1}{c}{\bf Methods} & \multicolumn{1}{c}{\bf Cora} & \multicolumn{1}{c}{\bf Citeseer} & \multicolumn{1}{c} {\bf Pubmed} &\multicolumn{1}{c}{\bf Actor} & \multicolumn{1}{c} {\bf Chameleon}& \multicolumn{1}{c}{\bf Squirrel} \\ \midrule
 \GCN   &	   86.98 $\pm$ 1.27  	&	   76.50 $\pm$ 1.36  	&	   88.42 $\pm$ 0.50    	&	   27.32 $\pm$ 1.10    	&	   64.82 $\pm$ 2.24     	&	   53.43 $\pm$ 2.01    	\\
\GCNII  &	\underline{88.37 $\pm$ 1.25}   & \underline{77.33 $\pm$ 1.48} & \underline{90.15 $\pm$ 0.43} & 37.44 $\pm$ 1.30 & 63.86 $\pm$ 3.04 & 38.47 $\pm$ 1.58   \\
\GAT    &	87.30 $\pm$ 1.10   &  76.55 $\pm$ 1.23 & 86.33 $\pm$ 0.48 & 27.44 $\pm$ 0.89 & 60.26 $\pm$ 2.50  & 40.72 $\pm$ 1.55  \\
\MixHop  &	87.61 $\pm$ 0.85   & 76.26 $\pm$ 1.33 & 85.31 $\pm$ 0.61 & 32.22 $\pm$ 2.34 &  60.50 $\pm$ 2.53 & 43.80 $\pm$ 1.48   \\
\HGCN   &	87.87 $\pm$ 1.20   & 77.11 $\pm$ 1.57 & 89.49 $\pm$ 0.38 & 35.70 $\pm$ 1.00 & 60.11 $\pm$ 2.15 & 36.48 $\pm$ 1.86   \\
\LINKX  &	84.64 $\pm$ 1.13   & 73.19 $\pm$ 0.99 & 87.86 $\pm$ 0.77 & 36.10 $\pm$ 1.55 & 68.42 $\pm$ 1.38 & 61.81 $\pm$ 1.80   \\
\revise{\WRGAT} & \revise{88.20 $\pm$ 2.26} & \revise{76.81 $\pm$ 1.89} & \revise{88.52 $\pm$ 0.92} & \revise{36.53 $\pm$ 0.77} & \revise{65.24 $\pm$ 0.87} & \revise{48.85 $\pm$ 0.78} \\
\ACM    &	87.91 $\pm$ 0.95   & 77.32 $\pm$ 1.70 & 90.00 $\pm$ 0.52 & 36.28 $\pm$ 1.09 & 66.93 $\pm$ 1.85  & 54.40 $\pm$ 1.88   \\
\GloGNN &	88.33 $\pm$ 1.09   & 77.22 $\pm$ 1.78 & 89.24 $\pm$ 0.39 & \underline{37.70 $\pm$ 1.40} & \underline{71.21 $\pm$ 1.84}  & \underline{57.88 $\pm$ 1.76}   \\
\ours   &	\revise{\bf{89.12 $\pm$ 0.87}}  & \revise{\bf{80.28 $\pm$ 1.31}} & \revise{\bf{90.19 $\pm$ 0.41}} & \revise{\bf{37.79 $\pm$ 1.11}} & \revise{\bf{73.66 $\pm$ 2.44}} & \revise{\bf{64.26 $\pm$ 1.46}}   \\ \bottomrule 
\end{tabular}}\vspace{-2mm}
\end{table}

Table~\ref{tbl:polyfilter} and Table~\ref{tbl:modelopt} present the results of \ours compared with existing polynomial filters and model-optimized methods for node classification respectively. For ease of exposition, we highlight the {\em highest} accuracy score in bold and underline the {\em second highest} score for each dataset. 

As shown, our method \ours consistently achieves the highest accuracy scores on both the homophily datasets and heterophily datasets, except in one case on Actor in Table~\ref{tbl:polyfilter}. \ours exhibits explicit performance advantages over both SOTA polynomial filter \Specformer and SOTA model-optimized method \GloGNN for the majority of cases. In particular, the performance improvements are remarkably significant on the two heterophily datasets Chameleon and Squirrel. Specifically, the corresponding performance gains reach up to \revise{$1.03\%$} and \revise{$2.76\%$} in Table~\ref{tbl:polyfilter} and \revise{$2.45\%$} and \revise{$6.38\%$} in Table~\ref{tbl:modelopt} respectively. It is worth mentioning that the computation time of \newbasis is linear to graph sizes and propagation hops. The superior performance of \ours strongly confirms the superb effectiveness and generality of \newbasis.

\subsection{Spectrum distribution of datasets}\label{sec:spectrum}

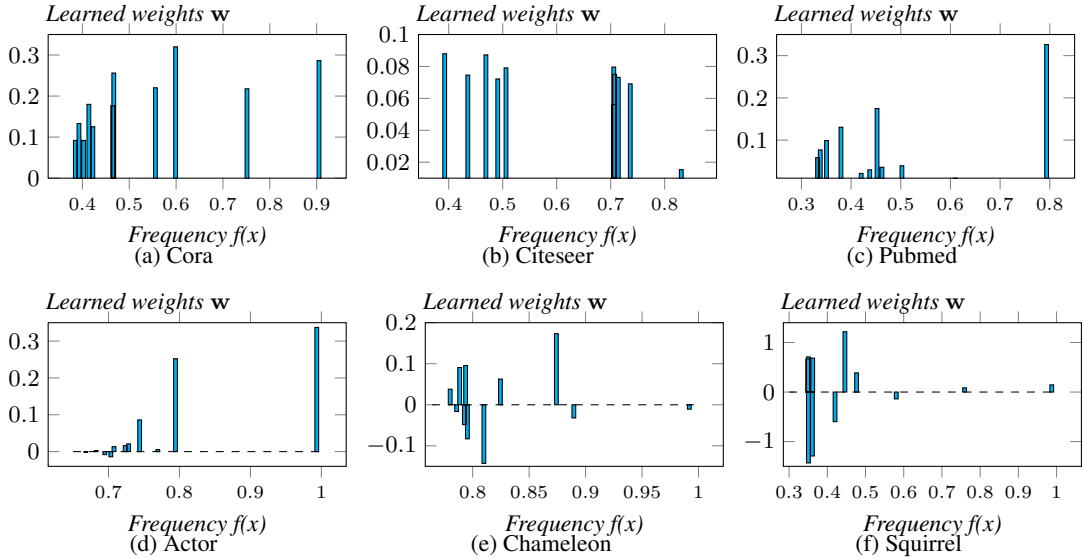
\begin{figure}
\centering
\begin{small}
\subfloat[{Cora}]{
\begin{tikzpicture}
\begin{axis}[
    ybar=2pt,
    height=\columnwidth/4,
    width=\columnwidth/2.52,            
    bar width=0.05cm,
    enlarge x limits=true,
    ymin=0, ymax=0.35,
    xmin=0.38, xmax=0.91,         
    xlabel={{\em Frequency f(x)}},
    ylabel={\em Learned weights $\w$},
    xtick={0.4,0.5,0.6,0.7,0.8,0.9},
    xticklabel style={font=\scriptsize},
    every axis y label/.style={at={(current axis.north west)},right=12mm,above=0mm},   
    ]
    \addplot[fill=cyan] coordinates {(0.9049213, 0.28665602) (0.7512426, 0.21770808)(0.5989015, 0.31996652)(0.55581844, 0.22022095)(0.46699378, 0.25615853)(0.46476635, 0.17635748)(0.4138301, 0.17979105)(0.42232922, 0.12524088)(0.39268094, 0.13313477)(0.40258533, 0.09199194)(0.38496265, 0.09186243)};    
\end{axis}
\end{tikzpicture}
\label{subfig:cora}
}
\subfloat[{Citeseer}]{
\begin{tikzpicture}
\begin{axis}[
    ybar=2pt,
    height=\columnwidth/4,
    width=\columnwidth/2.52,            
    bar width=0.05cm,
    enlarge x limits=true,
    ymin=0.01, ymax=0.1,
    xmin=0.39, xmax=0.85,
    xlabel={{\em Frequency f(x)}},
    ylabel={\em Learned weights $\w$},
    yticklabel={\pgfmathprintnumber[fixed,precision=2]{\tick}},
    xtick={0.4,0.5,0.6,0.7,0.8},
    xticklabel style={font=\scriptsize},
    every axis y label/.style={at={(current axis.north west)},right=12mm,above=0mm},   
    ]
    \addplot[fill=cyan] coordinates {(0.8306937, 0.015354116) (0.39249885, 0.08794039)(0.7361769, 0.06907876)(0.4353334, 0.074572735)(0.7140484, 0.07314714)(0.46884233, 0.08724001)(0.7073295, 0.075021446)(0.49079067, 0.07216103)(0.7056873, 0.0796282)(0.50604355, 0.079077415)(0.705987, 0.056096457)};    
\end{axis}
\end{tikzpicture}
\label{subfig:citeseer}
}
\subfloat[{Pubmed}]{
\begin{tikzpicture}
\begin{axis}[
    ybar=2pt,
    height=\columnwidth/4,
    width=\columnwidth/2.52,            
    bar width=0.05cm,
    enlarge x limits=true,
    ymin=0.01, ymax=0.350,
    xmin=0.3, xmax=0.8,         
    xlabel={{\em Frequency f(x)}},
    ylabel={\em Learned weights $\w$},
    xtick={0.3, 0.4,0.5,0.6,0.7,0.8},
    xticklabel style={font=\scriptsize},
    every axis y label/.style={at={(current axis.north west)},right=12mm,above=0mm},   
    ]
    \addplot[fill=cyan] coordinates {(0.79359525, 0.3262818)(0.60967284, 0.007923197)(0.45221677, 0.17476733)(0.5022794, 0.03927678)(0.37954324, 0.1305862)(0.46210948, 0.03577217)(0.3504402, 0.09895706)(0.43751678, 0.029734226)(0.3378904, 0.076884076)(0.4203169, 0.02118163)(0.3325249, 0.058505483)};    
\end{axis}
\end{tikzpicture}
\label{subfig:pubmed}
}
\vspace{-2mm}
\subfloat[{Actor}]{
\begin{tikzpicture}
\begin{axis}[
    ybar=2pt,
    height=\columnwidth/4,
    width=\columnwidth/2.52,            
    bar width=0.05cm,
    enlarge x limits=true,
    ymin=-0.04, ymax=0.35,
    xmin=0.65, xmax=1,         
    xlabel={{\em Frequency f(x)}},
    ylabel={\em Learned weights $\w$},
    xtick={0.7,0.8,0.9,1.0},
    xticklabel style={font=\scriptsize},
    every axis y label/.style={at={(current axis.north west)},right=12mm,above=0mm},]
    \addplot[fill=cyan] coordinates {(0.99351937, 0.3371407)(0.7944303, 0.25218594)(0.743898, 0.08607966)(0.7694793, 0.005307203)(0.6680614, -0.0017618219)(0.69484097, -0.008277266)(0.7032763, -0.014003986)(0.6826355, 0.0028506124)(0.72351223, 0.016178958)(0.7077483, 0.013457422)(0.7282677, 0.02094628)}; 
    \draw[black, dashed] ( 0.65, 0) -- ( 1, 0);    
\end{axis}
\end{tikzpicture}
\label{subfig:actor}
}
\subfloat[{Chameleon}]{
\begin{tikzpicture}
\begin{axis}[
    ybar=2pt,
    height=\columnwidth/4,
    width=\columnwidth/2.52,            
    bar width=0.05cm,
    enlarge x limits=true,
    ymin=-0.15, ymax=0.2,
    xmin=0.78, xmax=1.0,         
    xlabel={{\em Frequency f(x)}},
    ylabel={\em Learned weights $\w$},
    xtick={0.8,0.85, 0.9, 0.95, 1.0},
    xticklabel style={font=\scriptsize},
    every axis y label/.style={at={(current axis.north west)},right=12mm,above=0mm},   
    ]
    \addplot[fill=cyan] coordinates {(0.99189526, -0.010718918)(0.8896081, -0.032063093)(0.8741272, 0.17320552)(0.8245465, 0.0626653)(0.8099159, -0.1427136)(0.78005356, 0.03798501)(0.78844684, 0.09077762)(0.7926478, -0.04823897)(0.7953111, -0.08292176)(0.78577465, -0.016427025)(0.7937439, 0.095499545)};
    \draw[black, dashed] ( 0.75, 0) -- ( 1, 0);    
\end{axis}
\end{tikzpicture}
\label{subfig:chameleon}
}
\subfloat[{Squirrel}]{
\begin{tikzpicture}
\begin{axis}[
    ybar=2pt,
    height=\columnwidth/4,
    width=\columnwidth/2.52,            
    bar width=0.05cm,
    enlarge x limits=true,
    ymin=-1.5, ymax=1.4,
    xmin=0.35, xmax=1.0,         
    xlabel={{\em Frequency f(x)}},
    ylabel={\em Learned weights $\w$},
    xtick={0.3, 0.4, 0.5, 0.6, 0.7, 0.8, 0.9, 1.0},
    xticklabel style={font=\scriptsize},
    every axis y label/.style={at={(current axis.north west)},right=12mm,above=0mm},   
    ]
    \addplot[fill=cyan] coordinates {(0.9874364, 0.14482574)(0.75930494, 0.08507316)(0.44606453, 1.217764)(0.58042026, -0.13832489)(0.35057908, -1.4312787)(0.47658646, 0.38738832)(0.36176968, 0.68456066)(0.42023587, -0.5975129)(0.36092305, -1.290962)(0.35031483, 0.7090253)(0.3487408, 0.6671545)};
    \draw[black, dashed] ( 0.3, 0) -- ( 1, 0);    
\end{axis}
\end{tikzpicture}
\label{subfig:squirrel}
}
\end{small}
\captionof{figure}{Spectrum distribution of trained \newbasis.} \label{fig:datasetspectrum}  
\end{figure}

\eat{to uncover the underlying characteristics of graph structures and feature signals}

The superior performance of \ours explicitly implies the superb capability of \newbasis to capture the spectral characteristics of graphs. For better demonstration, we first calculate the {\em spectral signal frequency} of each basis vector for all $d$-dimensions, resulting in $d$ spectrum of length $K+1$. We then average the spectrum and associate it with the learned weights $\w$ accordingly, where weights $\w \in \R^{K+1}$ of \newbasis are trained for each dataset. The spectrum distributions of trained \newbasis for the $6$ datasets are plotted in Figure~\ref{fig:datasetspectrum}.

\revise{Recall that signals in specific frequencies with weights in large absolute are enhanced while signals with small weights are suppressed.} As displayed, the majority of signals of the three homophily datasets lie within the relatively low-frequency intervals as expected, \eg $[0.3,0.5]$. We also observe some minor high-frequency information which also provides insightful information for node classification~\citep{KlicperaWG19, Chen0XYKRYF19, balcilar2020bridging}. On the contrary, \newbasis of the three heterophily datasets intends to remove low-frequency signals with negative weights and preserve high-frequency information. The distinct spectrum distributions of \newbasis disclose the unique spectral characteristics of each dataset. Those results manifest the capability of \newbasis as a new method to analyze graphs with varying heterophily degrees in the spectral domain with enriched interpretability.

\subsection{Ablation Study}\label{sec:ablation}

\begin{figure}[!t]
  \centering
  \begin{minipage}[t]{0.48\textwidth}
    \centering
    \begin{tikzpicture}[scale=1,every mark/.append style={mark size=1.5pt}]
        \begin{axis}[
            height=\columnwidth/1.8,
            width=\columnwidth/1,
            ylabel={\em Accuracy gap} (\%),
            xlabel={\em Varying homo. ratio $h$},
            xmin=0.5, xmax=8.5,
            ymin=-0.2, ymax=4,
            xtick={1,2,3,4,5,6,7,8},
            xticklabel style={font=\scriptsize},
            yticklabel style={font=\footnotesize},
            xticklabels={0.13,0.2, 0.3, 0.4, 0.5, 0.6, 0.7, 0.81},
            every axis y label/.style={font=\footnotesize,at={(current axis.north west)},right=9mm,above=0mm},
            legend columns=3,
            legend style={fill=none,font=\footnotesize,at={(0.35,1.3)},anchor=center,draw=none},
        ]
        \addplot[line width=0.25mm,mark=triangle,color=cyan]
        coordinates {
            (1, 0) (2, 0) (3, 1.38) (4, 6.86) (5, 10.59) (6,17.76) (7,29.78) (8,35.76) }; 
        \addplot[line width=0.25mm,mark=diamond,color=blue]
        coordinates {
            (1, 1.6) (2, 0.43) (3,0) (4, 0) (5, 0.52) (6, 0.51) (7, 0.13) (8,0)}; 
        \addplot[line width=0.25mm,mark=o,color=red]
        coordinates {
            (1, 0.73) (2, 1.55) (3,2.69) (4, 1.79) (5, 1.48) (6, 1.4) (7, 0.15) (8, 0.71) }; 
        \legend{\HetFilter, \HomFilter, \OrtFilter}
        \end{axis}
    \end{tikzpicture}
    \caption{Accuracy gaps of the three variants from \ours on $\G_s$ across varying $h$.} \label{fig:varyingh}
  \end{minipage}
  \hspace{2mm}
  \begin{minipage}[t]{0.48\textwidth}
    \centering
   \begin{tikzpicture}[scale=1,every mark/.append style={mark size=1.5pt}]
    \begin{axis}[
        height=\columnwidth/1.8,
        width=\columnwidth/1,
        ylabel={\em Accuracy (\%)},
        xlabel={\em Varying $\tau$},
        xmin=0.5, xmax=10.5,
        ymin=55, ymax=92,
        xtick={1,2,3,4,5,6,7,8,9,10},
        xticklabel style = {font=\footnotesize},
        yticklabel style = {font=\footnotesize},
        xticklabels={ 0.1,0.2,0.3,0.4,0.5,0.6,0.7,0.8,0.9,1},
        every axis y label/.style={font=\footnotesize,at={(current axis.north west)},right=9mm,above=0mm},
        legend columns=2,
        legend style={fill=none,font=\footnotesize,at={(0.35,1.3)},anchor=center,draw=none},
    ] 
    \addplot [line width=0.25mm, color=blue, error bars/.cd, y dir=both, y explicit] 
    table [x=x, y=y, y error=y-err]{%
    x   y   y-err
    1	70.53	2.31
    2	72.02	1.34
    3	74.24	2.19
    4	77.54	0.88
    5	78.90	1.61
    6	81.87	1.36
    7	84.56	2.11
    8	86.93	1.04
    9	89.39	1.36
    10  89.11   1.48
    };
    \addplot [line width=0.25mm, color=red, error bars/.cd, y dir=both, y explicit] 
    table [x=x, y=y, y error=y-err]{%
    x   y   y-err
    1	66.75	1.05
    2	66.87	1.70
    3	65.85	1.51
    4	66.62	0.85
    5	66.66	0.96
    6	67.09	1.08
    7	67.01	1.25
    8	66.07	0.65
    9	65.40	1.39
    10  58.65   1.15
    };
    \legend{\textsf{Cora}, \textsf{Squirrel}}
    \end{axis}
\end{tikzpicture}
\caption{Accuracy (\%) with varying $\tau$.} \label{fig:varyingtau}
  \end{minipage}
\end{figure}

\spara{Universality of \newbasis} We compare \ours with three of its variants using distinct polynomial bases in order to verify the effectiveness of \newbasis. To this end, we alter \ours by changing \newbasis into 1) a filter simply using the heterophily basis (setting $\tau=0$) denoted as \HetFilter, 2) a filter simply using the homophily basis (setting $\tau=1$) denoted as \HomFilter, and 3) a filter using the orthonormal basis (adopting $\{\v_0, \v_1, \cdots, \v_k\}$) denoted as \OrtFilter. For easy control, we generate a synthetic dataset $\G_s$ by adopting the graph structure and label set of Cora. W.l.o.g., we generate a random one-hot feature vector in $100$ dimensions for each node in $\G_s$. To vary homophily ratios of $\G_s$, we permute nodes in a random sequence and randomly reassign node labels progressively, resulting in homophily ratios in $\{0.13\footnote{Note that this is the smallest homophily ratio we can possibly acquire by random reassignments.}, 0.20, 0.30, 0.40, 0.50, 0.60, 0.70, 0.81\}$ accordingly. 

The performance advantage gaps of \ours over the three variants are presented in Figure~\ref{fig:varyingh}. We omit the results of \HetFilter beyond $h\ge 0.3$ since the corresponding performance gaps become significantly larger, which is as expected since the heterophily basis is incapable of tackling homophily graphs. In particular, the performance advantage of \ours over \HomFilter gradually decreases as $h$ grows larger. Contrarily, performance gaps of \OrtFilter from \ours peak at $h=0.3$ with a notable shortfall and then erratically decrease, ending with an accuracy gap of $0.71\%$ at $h=0.81$. The fluctuation of \OrtFilter states the inferiority of orthonormal basis over \newbasis.

\spara{Sensitivity of $\tau$} To explore the sensitivity of \ours towards the hyperparameter $\tau$, we vary $\tau$ in $\{0,0.1,\cdots,0.9,1\}$ and test \ours on the strong homophily dataset Cora and the strong heterophily dataset Squirrel. Figure~\ref{fig:varyingtau} plots the performance development along varying $\tau$. As displayed, \ours prefers the homophily basis on Cora, and the performance peaks at $\tau=0.9$. On the contrary, the performance of \ours slightly fluctuates when $\tau \le 0.7$ and then moderately decreases along the increase of $\tau$ on Squirrel. When $\tau=1.0$, the accuracy score drops sharply since only the homophily basis is utilized in this scenario.

 \vspace{-2mm}

\section{Related Work}\label{sec:relatedwork}

\spara{Polynomial filters} As the seminal work, \ChebNet~\citep{DefferrardBV16} utilizes a $K$-order truncated Chebyshev polynomials~\citep{mason2002chebyshev, hammond2011wavelets} and provides $K$-hop localized filtering capability. \GPRGNN~\citep{ChienP0M21} simply adopts monomials instead and applies the generalized PageRank~\citep{0005CM19} scores as the coefficients to measure node proximity. In comparison, \SGC~\citep{WuSZFYW19} simplifies the propagation by keeping only the $K$-th order polynomial and removing nonlinearity. \revise{\ASGC~\cite{ChanpuriyaM22} simplifies the graph convolution operation by calculating a trainable Krylov matrix so as to adapt various heterophily graphs, which, however, is suboptimal as demonstrated in our experiments.} To enhance controllability and interpretability, \BernNet~\citep{he2021bernnet} employs nonnegative Bernstein polynomials as the basis. Later, \citet{WangZ22} examine the expressive power of existing polynomials and propose \JacobiConv by leveraging Jacobi polynomial~\citep{askey1974positive}, achieving better adaptability to underlying graphs. Subsequently, \citet{HeWW22} revisit \ChebNet and pinpoint the over-fitting issue in Chebyshev approximation. To address the issue, they turn to Chebyshev interpolation and propose \ChebNetII. Recently, polynomial filter~\OptBasisGNN~\citep{GuoW23} orthogonalizes the polynomial basis in order to maximize the convergence speed. Instead of using fixed-order polynomials, \Specformer~\citep{BoSWL23} resorts to Transformer~\citep{VaswaniSPUJGKP17} to derive learnable bases for each feature dimension. While \Specformer demonstrates promising performance, it requires conducting eigendecomposition with the cost of $O(n^3)$, rendering it impractical for large social graphs. Contrarily, the time complexity of \ours is {\em linear} to both graph sizes and propagation hops. Nonetheless, none of the above polynomial filters take the varying heterophily degrees of graphs into consideration when utilizing polynomials, which leads to suboptimal empirical performance, as verified in our experiments.

\spara{Model-optimized GNNs} One commonly adopted technique in model design is to combine both low-pass and high-pass filters. \textsf{GNN-LF/HF}~\citep{ZhuWSJ021} devises variants of the Laplacian matrix to construct a low-pass and high-pass filter respectively. \revise{HOG-GNN~\cite{WangJWHH22} designs a new propagation mechanism and considers the heterophily degrees between node pairs during neighbor aggregation, which is optimized from a spatial perspective. DualGR~\cite{LingC0P00023} focuses on the multi-view graph clustering problem and proposes dual label-guided graph refinement to handle heterophily graphs, which is a graph-level classification task.} \ACM~\citep{LuanHLZZZCP22} trains both low-pass and high-pass filters in each layer and then adopts the embeddings from each filter adaptively. Another applied model design aims to extract homophily from both local and global graph structures. \HGCN~\citep{ZhuYZHAK20} incorporates ego and neighbor embeddings, and high-order neighborhood and intermediate representations. Similarly, \GloGNN~\citep{LiZCSLLQ22} trains a coefficient matrix in each layer to measure the correlations between nodes so as to aggregate homophilous nodes globally. To explicitly capture the relations between distant nodes, \WRGAT~\citep{SureshBNLM21} leverages the graph rewiring~\citep{ToppingGC0B22, KarhadkarBM23} technique by constructing new edges with weights to measure node proximity. Additionally, there are GNNs handling heterophily graphs from other aspects. \LINKX~\citep{lim2021large} learns embeddings from node features and graph structure in parallel. Then the two embeddings are concatenated and fed into MLP for node predictions. \OrderedGNN~\citep{SongZWL23} establishes the hierarchy structure of neighbors and then constrains the neighbor nodes within specific hops into the specific blocks of neurons, avoiding feature mixing within hops.

\section{Conclusion}\label{sec:con}

In this paper, we propose a universal polynomial basis \newbasis by incorporating the graph heterophily degrees in the premise of thorough theoretical analysis for spectral graph neural networks. Upon \newbasis, we devise a general graph filter \ours. By a comprehensive evaluation of \ours on both real-world and synthetic datasets against a wide range of baselines, the remarkably superior performance of \ours significantly supports the effectiveness and generality of \newbasis for graphs with varying heterophily. In addition, \newbasis is exhibited as a promising new method for graph analysis by capturing the spectral characteristics of graphs and enriching the interpretability.

\clearpage
\bibliography{ref}

\begin{thebibliography}{49}
\providecommand{\natexlab}[1]{#1}
\providecommand{\url}[1]{\texttt{#1}}
\expandafter\ifx\csname urlstyle\endcsname\relax
  \providecommand{\doi}[1]{doi: #1}\else
  \providecommand{\doi}{doi: \begingroup \urlstyle{rm}\Url}\fi

\bibitem[Abu{-}El{-}Haija et~al.(2019)Abu{-}El{-}Haija, Perozzi, Kapoor,
  Alipourfard, Lerman, Harutyunyan, Steeg, and Galstyan]{AbuElHaijaPKA19}
Sami Abu{-}El{-}Haija, Bryan Perozzi, Amol Kapoor, Nazanin Alipourfard,
  Kristina Lerman, Hrayr Harutyunyan, Greg~Ver Steeg, and Aram Galstyan.
\newblock Mixhop: Higher-order graph convolutional architectures via sparsified
  neighborhood mixing.
\newblock In \emph{{ICML}}, volume~97, pp.\  21--29, 2019.

\bibitem[Askey(1974)]{askey1974positive}
Richard Askey.
\newblock Positive jacobi polynomial sums, iii.
\newblock In \emph{Linear Operators and Approximation II}, pp.\  305--312.
  1974.

\bibitem[Balcilar et~al.(2020)Balcilar, Renton, H{\'e}roux, Gauzere, Adam, and
  Honeine]{balcilar2020bridging}
Muhammet Balcilar, Guillaume Renton, Pierre H{\'e}roux, Benoit Gauzere,
  Sebastien Adam, and Paul Honeine.
\newblock Bridging the gap between spectral and spatial domains in graph neural
  networks.
\newblock \emph{arXiv preprint arXiv:2003.11702}, 2020.

\bibitem[Bo et~al.(2023)Bo, Shi, Wang, and Liao]{BoSWL23}
Deyu Bo, Chuan Shi, Lele Wang, and Renjie Liao.
\newblock Specformer: Spectral graph neural networks meet transformers.
\newblock In \emph{{ICLR}}, 2023.

\bibitem[Boyd et~al.(2004)Boyd, Boyd, and Vandenberghe]{boyd2004convex}
Stephen Boyd, Stephen~P Boyd, and Lieven Vandenberghe.
\newblock \emph{Convex optimization}.
\newblock Cambridge university press, 2004.

\bibitem[Chanpuriya \& Musco(2022)Chanpuriya and Musco]{ChanpuriyaM22}
Sudhanshu Chanpuriya and Cameron Musco.
\newblock Simplified graph convolution with heterophily.
\newblock In \emph{NeurIPS}, 2022.

\bibitem[Chen et~al.(2020)Chen, Wei, Huang, Ding, and Li]{ChenWHDL20}
Ming Chen, Zhewei Wei, Zengfeng Huang, Bolin Ding, and Yaliang Li.
\newblock Simple and deep graph convolutional networks.
\newblock In \emph{{ICML}}, volume 119, pp.\  1725--1735, 2020.

\bibitem[Chen et~al.(2019)Chen, Fan, Xu, Yan, Kalantidis, Rohrbach, Yan, and
  Feng]{Chen0XYKRYF19}
Yunpeng Chen, Haoqi Fan, Bing Xu, Zhicheng Yan, Yannis Kalantidis, Marcus
  Rohrbach, Shuicheng Yan, and Jiashi Feng.
\newblock Drop an octave: Reducing spatial redundancy in convolutional neural
  networks with octave convolution.
\newblock In \emph{{ICCV}}, pp.\  3434--3443, 2019.

\bibitem[Chien et~al.(2021)Chien, Peng, Li, and Milenkovic]{ChienP0M21}
Eli Chien, Jianhao Peng, Pan Li, and Olgica Milenkovic.
\newblock Adaptive universal generalized pagerank graph neural network.
\newblock In \emph{{ICLR}}, 2021.

\bibitem[Chung \& Graham(1997)Chung and Graham]{chung1997spectral}
Fan~RK Chung and Fan~Chung Graham.
\newblock \emph{Spectral graph theory}.
\newblock Number~92. American Mathematical Soc., 1997.

\bibitem[Defferrard et~al.(2016)Defferrard, Bresson, and
  Vandergheynst]{DefferrardBV16}
Micha{\"{e}}l Defferrard, Xavier Bresson, and Pierre Vandergheynst.
\newblock Convolutional neural networks on graphs with fast localized spectral
  filtering.
\newblock In \emph{NIPS}, pp.\  3837--3845, 2016.

\bibitem[Farouki(2012)]{Farouki12}
Rida~T. Farouki.
\newblock The bernstein polynomial basis: {A} centennial retrospective.
\newblock \emph{Comput. Aided Geom. Des.}, 29\penalty0 (6):\penalty0 379--419,
  2012.

\bibitem[Frasca et~al.(2020)Frasca, Rossi, Eynard, Chamberlain, Bronstein, and
  Monti]{frasca2020sign}
Fabrizio Frasca, Emanuele Rossi, Davide Eynard, Ben Chamberlain, Michael
  Bronstein, and Federico Monti.
\newblock Sign: Scalable inception graph neural networks.
\newblock \emph{arXiv preprint arXiv:2004.11198}, 2020.

\bibitem[Gautschi(2004)]{gautschi2004orthogonal}
Walter Gautschi.
\newblock \emph{Orthogonal polynomials: computation and approximation}.
\newblock OUP Oxford, 2004.

\bibitem[Guo \& Wei(2023)Guo and Wei]{GuoW23}
Yuhe Guo and Zhewei Wei.
\newblock Graph neural networks with learnable and optimal polynomial bases.
\newblock In \emph{{ICML}}, volume 202, pp.\  12077--12097, 2023.

\bibitem[Hamilton et~al.(2017)Hamilton, Ying, and Leskovec]{HamiltonYL17}
William~L. Hamilton, Zhitao Ying, and Jure Leskovec.
\newblock Inductive representation learning on large graphs.
\newblock In \emph{{Neurips}}, pp.\  1024--1034, 2017.

\bibitem[Hammond et~al.(2011)Hammond, Vandergheynst, and
  Gribonval]{hammond2011wavelets}
David~K Hammond, Pierre Vandergheynst, and R{\'e}mi Gribonval.
\newblock Wavelets on graphs via spectral graph theory.
\newblock \emph{Applied and Computational Harmonic Analysis}, 30\penalty0
  (2):\penalty0 129--150, 2011.

\bibitem[He et~al.(2021)He, Wei, Xu, et~al.]{he2021bernnet}
Mingguo He, Zhewei Wei, Hongteng Xu, et~al.
\newblock Bernnet: Learning arbitrary graph spectral filters via bernstein
  approximation.
\newblock 2021.

\bibitem[He et~al.(2022)He, Wei, and Wen]{HeWW22}
Mingguo He, Zhewei Wei, and Ji{-}Rong Wen.
\newblock Convolutional neural networks on graphs with chebyshev approximation,
  revisited.
\newblock In \emph{NeurIPS}, 2022.

\bibitem[Karhadkar et~al.(2023)Karhadkar, Banerjee, and
  Mont{\'{u}}far]{KarhadkarBM23}
Kedar Karhadkar, Pradeep~Kr. Banerjee, and Guido Mont{\'{u}}far.
\newblock Fosr: First-order spectral rewiring for addressing oversquashing in
  gnns.
\newblock In \emph{{ICLR}}, 2023.

\bibitem[Kingma \& Ba(2015)Kingma and Ba]{KingmaB14}
Diederik~P. Kingma and Jimmy Ba.
\newblock Adam: {A} method for stochastic optimization.
\newblock In \emph{{ICLR}}, 2015.

\bibitem[Kipf \& Welling(2017)Kipf and Welling]{KipfW17}
Thomas~N. Kipf and Max Welling.
\newblock Semi-supervised classification with graph convolutional networks.
\newblock In \emph{{ICLR}}, 2017.

\bibitem[Klicpera et~al.(2019{\natexlab{a}})Klicpera, Bojchevski, and
  G{\"{u}}nnemann]{KlicperaBG19}
Johannes Klicpera, Aleksandar Bojchevski, and Stephan G{\"{u}}nnemann.
\newblock Predict then propagate: Graph neural networks meet personalized
  pagerank.
\newblock In \emph{{ICLR}}, 2019{\natexlab{a}}.

\bibitem[Klicpera et~al.(2019{\natexlab{b}})Klicpera, Wei{\ss}enberger, and
  G{\"{u}}nnemann]{KlicperaWG19}
Johannes Klicpera, Stefan Wei{\ss}enberger, and Stephan G{\"{u}}nnemann.
\newblock Diffusion improves graph learning.
\newblock In \emph{NeurIPS}, pp.\  13333--13345, 2019{\natexlab{b}}.

\bibitem[Lei et~al.(2022)Lei, Wang, Li, Ding, and Wei]{LeiWLDW22}
Runlin Lei, Zhen Wang, Yaliang Li, Bolin Ding, and Zhewei Wei.
\newblock Evennet: Ignoring odd-hop neighbors improves robustness of graph
  neural networks.
\newblock In \emph{NeurIPS}, 2022.

\bibitem[Li et~al.(2019)Li, Chien, and Milenkovic]{0005CM19}
Pan Li, I~(Eli) Chien, and Olgica Milenkovic.
\newblock Optimizing generalized pagerank methods for seed-expansion community
  detection.
\newblock In \emph{NeurIPS}, pp.\  11705--11716, 2019.

\bibitem[Li et~al.(2022)Li, Zhu, Cheng, Shan, Luo, Li, and Qian]{LiZCSLLQ22}
Xiang Li, Renyu Zhu, Yao Cheng, Caihua Shan, Siqiang Luo, Dongsheng Li, and
  Weining Qian.
\newblock Finding global homophily in graph neural networks when meeting
  heterophily.
\newblock In \emph{{ICML}}, volume 162, pp.\  13242--13256, 2022.

\bibitem[Liesen \& Strakos(2013)Liesen and Strakos]{liesen2013krylov}
J{\"o}rg Liesen and Zdenek Strakos.
\newblock \emph{Krylov subspace methods: principles and analysis}.
\newblock Oxford University Press, 2013.

\bibitem[Lim et~al.(2021)Lim, Hohne, Li, Huang, Gupta, Bhalerao, and
  Lim]{lim2021large}
Derek Lim, Felix Hohne, Xiuyu Li, Sijia~Linda Huang, Vaishnavi Gupta, Omkar
  Bhalerao, and Ser~Nam Lim.
\newblock Large scale learning on non-homophilous graphs: New benchmarks and
  strong simple methods.
\newblock \emph{NIPS}, 34:\penalty0 20887--20902, 2021.

\bibitem[Ling et~al.(2023)Ling, Chen, Ren, Pu, Xu, Zhu, and He]{LingC0P00023}
Yawen Ling, Jianpeng Chen, Yazhou Ren, Xiaorong Pu, Jie Xu, Xiaofeng Zhu, and
  Lifang He.
\newblock Dual label-guided graph refinement for multi-view graph clustering.
\newblock In \emph{{AAAI}}, pp.\  8791--8798, 2023.

\bibitem[Luan et~al.(2021)Luan, Hua, Lu, Zhu, Zhao, Zhang, Chang, and
  Precup]{luan2021heterophily}
Sitao Luan, Chenqing Hua, Qincheng Lu, Jiaqi Zhu, Mingde Zhao, Shuyuan Zhang,
  Xiao-Wen Chang, and Doina Precup.
\newblock Is heterophily a real nightmare for graph neural networks to do node
  classification?
\newblock \emph{arXiv preprint arXiv:2109.05641}, 2021.

\bibitem[Luan et~al.(2022)Luan, Hua, Lu, Zhu, Zhao, Zhang, Chang, and
  Precup]{LuanHLZZZCP22}
Sitao Luan, Chenqing Hua, Qincheng Lu, Jiaqi Zhu, Mingde Zhao, Shuyuan Zhang,
  Xiao{-}Wen Chang, and Doina Precup.
\newblock Revisiting heterophily for graph neural networks.
\newblock In \emph{NeurIPS}, 2022.

\bibitem[Mason \& Handscomb(2002)Mason and Handscomb]{mason2002chebyshev}
John~C Mason and David~C Handscomb.
\newblock \emph{Chebyshev polynomials}.
\newblock Chapman and Hall/CRC, 2002.

\bibitem[Pei et~al.(2020)Pei, Wei, Chang, Lei, and Yang]{PeiWCLY20}
Hongbin Pei, Bingzhe Wei, Kevin~Chen{-}Chuan Chang, Yu~Lei, and Bo~Yang.
\newblock Geom-gcn: Geometric graph convolutional networks.
\newblock In \emph{{ICLR}}, 2020.

\bibitem[Platonov et~al.(2022)Platonov, Kuznedelev, Babenko, and
  Prokhorenkova]{platonov2022}
Oleg Platonov, Denis Kuznedelev, Artem Babenko, and Liudmila Prokhorenkova.
\newblock Characterizing graph datasets for node classification: Beyond
  homophily-heterophily dichotomy.
\newblock \emph{arXiv preprint arXiv:2209.06177}, 2022.

\bibitem[Sen et~al.(2008)Sen, Namata, Bilgic, Getoor, Galligher, and
  Eliassi-Rad]{sen2008collective}
Prithviraj Sen, Galileo Namata, Mustafa Bilgic, Lise Getoor, Brian Galligher,
  and Tina Eliassi-Rad.
\newblock Collective classification in network data.
\newblock \emph{AI magazine}, 29\penalty0 (3):\penalty0 93--93, 2008.

\bibitem[Song et~al.(2023)Song, Zhou, Wang, and Lin]{SongZWL23}
Yunchong Song, Chenghu Zhou, Xinbing Wang, and Zhouhan Lin.
\newblock Ordered {GNN:} ordering message passing to deal with heterophily and
  over-smoothing.
\newblock In \emph{{ICLR}}, 2023.

\bibitem[Suresh et~al.(2021)Suresh, Budde, Neville, Li, and Ma]{SureshBNLM21}
Susheel Suresh, Vinith Budde, Jennifer Neville, Pan Li, and Jianzhu Ma.
\newblock Breaking the limit of graph neural networks by improving the
  assortativity of graphs with local mixing patterns.
\newblock In \emph{{KDD}}, pp.\  1541--1551, 2021.

\bibitem[Tang et~al.(2009)Tang, Sun, Wang, and Yang]{TangSWY09}
Jie Tang, Jimeng Sun, Chi Wang, and Zi~Yang.
\newblock Social influence analysis in large-scale networks.
\newblock In \emph{{SIGKDD}}, pp.\  807--816. {ACM}, 2009.

\bibitem[Topping et~al.(2022)Topping, Giovanni, Chamberlain, Dong, and
  Bronstein]{ToppingGC0B22}
Jake Topping, Francesco~Di Giovanni, Benjamin~Paul Chamberlain, Xiaowen Dong,
  and Michael~M. Bronstein.
\newblock Understanding over-squashing and bottlenecks on graphs via curvature.
\newblock In \emph{{ICLR}}, 2022.

\bibitem[Vaswani et~al.(2017)Vaswani, Shazeer, Parmar, Uszkoreit, Jones, Gomez,
  Kaiser, and Polosukhin]{VaswaniSPUJGKP17}
Ashish Vaswani, Noam Shazeer, Niki Parmar, Jakob Uszkoreit, Llion Jones,
  Aidan~N. Gomez, Lukasz Kaiser, and Illia Polosukhin.
\newblock Attention is all you need.
\newblock In \emph{NeurIPS}, pp.\  5998--6008, 2017.

\bibitem[Velickovic et~al.(2018)Velickovic, Cucurull, Casanova, Romero,
  Li{\`{o}}, and Bengio]{VelickovicCCRLB18}
Petar Velickovic, Guillem Cucurull, Arantxa Casanova, Adriana Romero, Pietro
  Li{\`{o}}, and Yoshua Bengio.
\newblock Graph attention networks.
\newblock In \emph{{ICLR}}, 2018.

\bibitem[Wang et~al.(2022)Wang, Jin, Wang, He, and Huang]{WangJWHH22}
Tao Wang, Di~Jin, Rui Wang, Dongxiao He, and Yuxiao Huang.
\newblock Powerful graph convolutional networks with adaptive propagation
  mechanism for homophily and heterophily.
\newblock In \emph{{AAAI}}, pp.\  4210--4218, 2022.

\bibitem[Wang \& Zhang(2022)Wang and Zhang]{WangZ22}
Xiyuan Wang and Muhan Zhang.
\newblock How powerful are spectral graph neural networks.
\newblock In \emph{{ICML}}, volume 162, pp.\  23341--23362, 2022.

\bibitem[Wright et~al.(1999)Wright, Nocedal, et~al.]{wright1999numerical}
Stephen Wright, Jorge Nocedal, et~al.
\newblock Numerical optimization.
\newblock \emph{Springer Science}, 35\penalty0 (67-68):\penalty0 7, 1999.

\bibitem[Wu et~al.(2019)Wu, Jr., Zhang, Fifty, Yu, and Weinberger]{WuSZFYW19}
Felix Wu, Amauri H.~Souza Jr., Tianyi Zhang, Christopher Fifty, Tao Yu, and
  Kilian~Q. Weinberger.
\newblock Simplifying graph convolutional networks.
\newblock In \emph{{ICML}}, volume~97, pp.\  6861--6871, 2019.

\bibitem[Zhou et~al.(2021)Zhou, Huang, Zha, Chen, Li, Choi, and
  Hu]{ZhouHZCLCH21}
Kaixiong Zhou, Xiao Huang, Daochen Zha, Rui Chen, Li~Li, Soo{-}Hyun Choi, and
  Xia Hu.
\newblock Dirichlet energy constrained learning for deep graph neural networks.
\newblock In \emph{NeurIPS}, pp.\  21834--21846, 2021.

\bibitem[Zhu et~al.(2020)Zhu, Yan, Zhao, Heimann, Akoglu, and
  Koutra]{ZhuYZHAK20}
Jiong Zhu, Yujun Yan, Lingxiao Zhao, Mark Heimann, Leman Akoglu, and Danai
  Koutra.
\newblock Beyond homophily in graph neural networks: Current limitations and
  effective designs.
\newblock In \emph{NeurIPS}, 2020.

\bibitem[Zhu et~al.(2021)Zhu, Wang, Shi, Ji, and Cui]{ZhuWSJ021}
Meiqi Zhu, Xiao Wang, Chuan Shi, Houye Ji, and Peng Cui.
\newblock Interpreting and unifying graph neural networks with an optimization
  framework.
\newblock In \emph{{WWW}}, pp.\  1215--1226, 2021.

\end{thebibliography}
\bibliographystyle{iclr2024}

\clearpage
\appendix
\section{Appendix}\label{sec:app}

\subsection{Proofs}\label{app:proofs}

\begin{proof}[Proof of Lemma~\ref{lem:frequencybound}]
Given any normalized signal $\x \in \R^n$ on graph $\G=(\V,\E)$, $f(\x)=\tfrac{\x^\top\L\x}{2}=\tfrac{\sum_{{\langle u,v\rangle}\in \E}(\x_u-\x_v)^2}{2\sum_{u\in \V}\x^2_u d_u}$. It is straightforward to infer $f(\x)\ge 0$. Meanwhile, it is known that $\sup_{\x \in \R^n}\tfrac{\sum_{{\langle u,v\rangle}\in \E}(\x_u-\x_v)^2}{\sum_{u\in \V}\x^2_u d_u} \le 2$~\citep{chung1997spectral}. Therefore $f(\x) \le 1$, which completes the proof. 
\end{proof}

\begin{proof}[Proof of Theorem~\ref{thm:frequencyratio}]
Let $\z=\mathrm{F(\w)}\x=\sum^K_{k=0}\w_k\P^k\x$. Then frequency $f(\textstyle\sum^K_{k=0}\w_k\P^k\x) =f(\z) =\tfrac{\z^\top \L\z}{2}=\tfrac{\sum_{{\langle u,v\rangle}\in \E}(\z_u-\z_v)^2}{2\sum_{u\in \V}\z^2_u d_u}$. It is noteworthy that $\mathrm{F(\w)}$ represents the optimal filter for node classification. This implies that node representations acquired by $\mathrm{F(\w)}$ exhibit similarity among nodes belonging to the same classes and dissimilarity among nodes belonging to distinct classes.

W.l.o.g, for $\forall u,v \in \V$, we assume a constant $\delta$ that $|\z_u - \z_v| \le c\delta$ with $c \ll 1$ if $\Y_u=\Y_v$; otherwise $|\z_u - \z_v|=g(\Y_u, \Y_v)\delta$ where $\Y\in \mathbb{N}^{n\times |\C|}$ is the one-hot label matrix and $g(\Y_u, \Y_v)\ge 1$ is a constant determined by $\Y_u$ and $\Y_v$. As a result, we have $\tfrac{\sum_{{\langle u,v\rangle}\in \E}(\z_u-\z_v)^2}{2\sum_{u\in \V}\z^2_u d_u}$ approaches to $\tfrac{c^2\delta^2 hm+\sum_{{\langle u,v\rangle}\in \E, \Y_u \neq \Y_v}g^2(\Y_u, \Y_v)\delta^2}{2\sum_{u\in \V}\z^2_u d_u}$. Since $c\ll 1$ and $g(\Y_u, \Y_v) \ge 1$, therefore $g^2(\Y_u, \Y_v) \gg c^2$ holds. In this regard, frequency $f(\textstyle\sum^K_{k=0}\w_k\P^k\x)$ is dominated by $\tfrac{\sum_{{\langle u,v\rangle}\in \E, \Y_u \neq \Y_v}g^2(\Y_u, \Y_v)\delta^2}{2\sum_{u\in \V}\z^2_u d_u}$, \ie proportional to $1-h$.
\end{proof}

\begin{proof}[Proof of Theorem~\ref{thm:homo}]
Consider a propagation matrix $\P$ and graph signal $\x$. Let $\{\lambda_1,\lambda_2,\cdots, \lambda_n\}$\footnote{For notation simplicity, we here reuse the notation $\lambda_i$ as the $i$-th eigenvalue of propagation matrix $\P$.} be the eigenvalues of $\P$ associated with eigenvectors $\{\v_1,\v_2,\cdots, \v_n\}$. For a general (non-bipartite) connected graph $\G$, we have $-1<\lambda_1 \le \lambda_2 \le \cdots \le \lambda_n =1$ and $\v^\top_i \v_j =0$ for $i \neq j$ and $\v^\top_i \v_j =1$ for $i = j$. In particular, we have $\lambda_n=1$ and $\v_n =\tfrac{\D^{\tfrac{1}{2}}\mathbf{1}}{\sqrt{2m}}$ where $\mathbf{1}\in \R^n$ is the all-one vector. Hence, $\P^k\x$ can be formulated as $\P^k\x=\sum^n_{i=1}\lambda^k_i(\v^\top_i\x)\v_i$. Specifically, we have \[\|\P^k\x\|=\sqrt{\left(\sum^n_{i=1}\lambda^k_i(\v^\top_i\x)\v_i \right)\cdot \left(\sum^n_{j=1}\lambda^k_j(\v^\top_j\x)\v_j \right)}=\sqrt{\sum^n_{i=1}\lambda^{2k}_i(\v^\top_i\x)^2}.\] Similarly, we have
$\P^k\x \cdot \P^{k+1}\x =\left(\sum^n_{i=1}\lambda^k_i(\v^\top_i\x)\right)\cdot \left(\sum^n_{i=1}\lambda^{k+1}_i(\v^\top_i\x)\right)=\sum^n_{i=1}\lambda^{2k+1}_i(\v^\top_i\x)^2$.  

Therefore, we have $\tfrac{\P^k\x \cdot \P^{k+1}\x}{\|\P^k\x\| \|\P^{k+1}\x\|}=\tfrac{\sum^n_{i=1}\lambda^{2k+1}_i(\v^\top_i\x)^2}{\sqrt{\sum^n_{i=1}\lambda^{2k}_i(\v^\top_i\x)^2}\sqrt{\sum^n_{i=1}\lambda^{2k+2}_i(\v^\top_i\x)^2}}$. For ease of exposition, we denote $c_k=\|\P^k\x\|$ and let $t$ be the integer index such that $\lambda_{t}<0\le \lambda_{t+1}$. As a result, we have 
\begin{equation}\label{eqn:unitvector}
	\sum^n_{i=1}\left(\tfrac{\lambda^k_i(\v^\top_i\x)}{c_k}\right)^2=\sum^n_{i=1}\left(\tfrac{\lambda^{k+1}_i(\v^\top_i\x)}{c_{k+1}}\right)^2=1    
\end{equation} and $\tfrac{\P^k\x \cdot \P^{k+1}\x}{\|\P^k\x\| \|\P^{k+1}\x\|}=\sum^t_{i=1}\tfrac{\lambda^{2k+1}_i(\v^\top_i\x)^2}{c_kc_{k+1}}+\sum^n_{i=t}\tfrac{\lambda^{2k+1}_i(\v^\top_i\x)^2}{c_kc_{k+1}}$. Note that the exponent $2k+1$ remains odd integer for any $k$ values. W.l.o.g., we denote the negative part as function $f_N(t,k)=\sum^t_{i=1}\tfrac{\lambda^{2k+1}_i(\v^\top_i\x)^2}{c_kc_{k+1}}$ and the positive part as function $f_P(t,k)=\sum^n_{i=t}\tfrac{\lambda^{2k+1}_i(\v^\top_i\x)^2}{c_kc_{k+1}}.$

Notice that $\lambda_i\in(-1,0)$ for $i\in \{1,\cdots,t\}$, $\lambda_i\in[0,1]$ ($\lambda_n=1$) for $i\in\{t+1,\cdots,n\}$, and $(\v^\top_i\x)^2$ for $i\in\{1,\cdots,n\}$ are constants. When $k$ increases, numerator $\sum^t_{i=1}\lambda^{2k}_i(\v^\top_i\x)^2$ of $\tfrac{\sum^t_{i=1} \lambda^{2k}_i(\v^\top_i\x)^2}{c^2_k}$ monotonically decreases and asymptotically approaches to $0$. According to Equation~\eqref{eqn:unitvector}, $\sum^t_{i=1}\tfrac{\lambda^{2k}_i(\v^\top_i\x)^2}{c^2_k}+\sum^n_{i=t}\tfrac{\lambda^{2k}_i(\v^\top_i\x)^2}{c^2_k}=1$ holds for all $k$. Therefore, when the case $\sum^t_{i=1}\lambda^{2k}_i(\v^\top_i\x)^2 > \sum^n_{i=t}\lambda^{2k}_i(\v^\top_i\x)^2$ occurs for certain $k$, there exists an integer $\eta \in \mathbb{N}$ such that $\sum^t_{i=1}\lambda^{2(\eta-1)}_i(\v^\top_i\x)^2 > \sum^n_{i=t}\lambda^{2(\eta-1)}_i(\v^\top_i\x)^2$ and $\sum^t_{i=1}\lambda^{2\eta}_i(\v^\top_i\x)^2 \le \sum^n_{i=t}\lambda^{2\eta}_i(\v^\top_i\x)^2.$ Considering the monotonicity of both $\sum^t_{i=1}\lambda^{2k}_i(\v^\top_i\x)^2$ and $\sum^n_{i=t}\lambda^{2k}_i(\v^\top_i\x)^2$, when $k\ge \eta$, $f_N(t,k)$ monotonically decreases while $f_P(t,k)$ monotonically increases with $k$. As a consequence, $\tfrac{\P^k\x \cdot \P^{k+1}\x}{\|\P^k\x\| \|\P^{k+1}\x\|}$ is monotonically increasing with $k$, and thus the angle angle $\arccos\left(\tfrac{\P^k\x \cdot \P^{k+1}\x}{\|\P^k\x\| \|\P^{k+1}\x\|}\right)$ is progressively smaller. \eat{Comments: The further explanation on why $f_N(t,k)$ monotonically decreases and $f_P(t,k)$ monotonically increases is as follows. Once observed $\sum^t_{i=1}\lambda^{2k}_i(\v^\top_i\x)^2 \le \sum^n_{i=t}\lambda^{2k}_i(\v^\top_i\x)^2$, their gap would increases since each component within the two expressions are with monotonicity. Being divided by the same factor $c^2_k$ does not change the ordering. Therefore, the positive part would be larger than the negative part, leading to an increase in cos value.}

Meanwhile, when $K\to \infty$, we have $\lim_{K\to\infty}\P^K\x=\lambda^K_n (\v^\top_n\x) \v_n=\tfrac{\v^\top_n\x}{\sqrt{2m}}\D^{\tfrac{1}{2}}\mathbf{1}$ where $\v_n =\tfrac{\D^{\tfrac{1}{2}}\mathbf{1}}{\sqrt{2m}}$. Therefore, $\lim_{K\to\infty}\arccos\left(\frac{\P^K\x \cdot \P^{K+1}\x}{\|\P^K\x\|\|\P^{K+1}\x\|}\right) \to 0 $ holds. 
\end{proof}

\begin{proof}[Proof of Theorem~\ref{thm:pivot}]
W.l.o.g, we consider a $k$-regular graph $\G=(\V, \E)$ with $n$ nodes. Given a random normalized signal $\x=(\x_1, \x_2, \cdots, \x_n)^\top$, $\phi \cdot \x =\sum^n_{i=1}\tfrac{\x_i}{\sqrt{n}}$. Meanwhile, $f(\x)=\tfrac{\x^\top\L\x}{2}=\tfrac{\sum_{{\langle u,v\rangle}\in \E}(\x_u-\x_v)^2}{2\sum_{u\in \V}\x^2_u d_u}=\tfrac{\sum_{{\langle u,v\rangle}\in \E}(\x_u-\x_v)^2}{2k}$. Over the randomness of $\G$, the expectation of the spectral signal frequency 
\begin{align*}
\mathbb{E}_{\G\sim \mathcal{G}}[f(\x)]
& = \tfrac{1}{2k}\cdot \tfrac{k}{n-1}\sum_{\forall u,v\in \V}\tfrac{(\x_u-\x_v)^2}{2} \\
& = \tfrac{1}{4(n-1)}\left(\sum_{u\in \V}(n-1)\x^2_u-\sum_{u\in \V}2\x_u\big(\sum_{v\in \V\setminus \{u\}}\x_v\big)\right) \\
& = \tfrac{1}{4(n-1)}\left(n-1-\sum_{u\in \V}2\x_u\big(\sum^n_{i=1}\x_i-\x_u\big)\right)\\
& = \tfrac{1}{4(n-1)}\left(n-1-2\big(\sum^n_{i=1}\x_i\big)^2+2 \right)\\
& = \tfrac{1}{4(n-1)}\left(n+1-2\big(\sum^n_{i=1}\x_i\big)^2\right) \\
& = \tfrac{n+1}{4(n-1)}-\tfrac{1}{2(n-1)}\big(\sum^n_{i=1}\tfrac{\x_i}{\sqrt{n}}\big)^2 \\
& = \tfrac{n+1-2(\phi \cdot \x)^2}{4(n-1)}
\end{align*}
As a consequence, if the angle $\theta:=\mathrm{arccos}( \phi \cdot \x )$ increases, $\phi\cdot \x$ decreases, resulting the increment of $\mathbb{E}_{\G\sim \mathcal{G}}[f(\x)]$, which completes the proof.
\end{proof}

Before the proof of Theorem~\ref{thm:heteroproperty}, we first introduce the following Lemma.
\begin{lemma}[Proposition 4.3~\citep{GuoW23}]~\label{lem:orthogonal} Vector $\v_k$ in Algorithm~\ref{alg:basis} is only dependent with $\v_{k-1}$ and $\v_{k-2}$.
\end{lemma}

\begin{proof}[Proof of Theorem~\ref{thm:heteroproperty}]
Based on Lemma~\ref{lem:orthogonal}, it is easy to prove that $\{\v_0, \v_1, \cdots,\v_K\}$ forms an orthonormal basis. Before proceeding, we first demonstrate that $\v_{k+1}$ is orthogonal to $\{\u_0,\u_1, \cdots, \u_k\}$ for $k\in \{0,1,\cdots, K\!-\!1\}$. In particular, we have $\u_0 = \v_0$ and $\u_k=\tfrac{\s_{k-1}/k+t_k\v_k}{\|\s_{k-1}/k+t_k\v_k\|}=\tfrac{\frac{1}{k}\sum^{k-1}_{i=0}\u_i+t_k\v_k}{\|\frac{1}{k}\sum^{k-1}_{i=0}\u_i+t_k\v_k\|}$ for $k\in [K]$ according to Algorithm~\ref{alg:basis}. Therefore, it is intuitive that there exist constants $\alpha_0, \alpha_1, \cdots, \alpha_k$ such that $\u_k=\sum^k_{i=0}\alpha_i \v_i$ holds. Since $\{\v_0, \v_1, \cdots,\v_K\}$ is an orthonormal basis, therefore $\v_{k+1}$ is orthogonal to $\{\u_0,\u_1, \cdots, \u_k\}$.

Denote $\theta:=\tfrac{(1-h)\pi}{2}$. First, we prove that $\u_0 \cdot \u_1 = \mathrm{cos}(\theta)$. In particular, we have $\u_1=\tfrac{\u_0+t_1\v_1}{\|\u_0+t_1\v_1\|}$ and $t_1=\sqrt{(\tfrac{\s_0^\top \u_0}{\mathrm{cos}(\theta)})^2-1}=\sqrt{\tfrac{1}{\mathrm{cos}^2(\theta)}-1}$. Then $\|\u_0+t_1\v_1\|=\sqrt{1+t^2_1}=\tfrac{1}{\mathrm{cos}(\theta)}$. Hence, $\u_0 \cdot \u_1 = \u^\top_0(\u_0+t_1\v_1)\mathrm{cos}(\theta)=\mathrm{cos}(\theta)$. 

Second, we assume that $\u_i \cdot \u_j=\mathrm{cos}(\theta)$ holds for $\forall i,j\in\{0,1,\cdots, k\!-\!1\}$ and $i\neq j$. In what follows, we then prove that $\u_k \cdot \u_j=\mathrm{cos}(\theta)$ holds for $j\in\{0,1,\cdots, k\!-\!1\}$. Specifically, we have $\u_k=\tfrac{\frac{1}{k}\sum^{k-1}_{i=0}\u_i+t_k\v_k}{\|\frac{1}{k}\sum^{k-1}_{i=0}\u_i+t_k\v_k\|}$. In particular, for the denominator, we have
\begin{align*}
\left\|\frac{1}{k}\sum^{k-1}_{i=0}\u_i+t_k\v_k\right\| 
& = \sqrt{\left(\frac{1}{k}\sum^{k-1}_{i=0}\u^\top_i+t_k\v^\top_k\right)\left(\frac{1}{k}\sum^{k-1}_{i=0}\u_i+t_k\v_k\right)} \\
& = \sqrt{\tfrac{\sum^{k-1}_{i=0}\u^\top_i \u_i+2\sum^{k-2}_{i=0}\u^\top_i(\sum^{k-1}_{j=i+1}\u_j)}{k^2}+t^2_k} \\
& = \sqrt{\tfrac{k+k(k-1)\mathrm{cos}(\theta)}{k^2}+\left(\tfrac{\s_{k-1}^\top\u_{k-1}}{k\mathrm{cos}(\theta)}\right)^2-\tfrac{(k-1)\mathrm{cos(\theta)}+1}{k}} \\
& = \frac{\sum^{k-1}_{i=0}\u^\top_i \cdot \u_{k-1} }{k\mathrm{cos}(\theta)} \\
& = \tfrac{1+(k-1)\mathrm{cos}(\theta)}{k\mathrm{cos}(\theta)}
\end{align*}
Meanwhile, we have $\u_k \cdot \u_j  = \tfrac{\frac{1}{k}\sum^{k-1}_{i=0}\u^\top_i \u_j+t_k\v^\top_k \u_j}{\|\frac{1}{k}\sum^{k-1}_{i=0}\u_i+t_k\v_k\|} = \tfrac{\frac{1}{k} (1+(k-1)\mathrm{cos}(\theta))\cdot k\mathrm{cos}(\theta)}{1+(k-1)\mathrm{cos}(\theta)}= \mathrm{cos}(\theta)$. 

Eventually, it is easy to verify that the derivation holds for $\forall j\in\{0,1,\cdots, k\!-\!1\}$, which completes the proof.
\end{proof}

\subsection{Experimental Settings}\label{app:settings}

\begin{table}[!t]
\centering
 \caption{Dataset Details.} \label{tbl:dataset}\vspace{-1mm} 
    \setlength{\tabcolsep}{0.5em}
 \small
\begin{tabular} {@{}l|rrrrrr@{}}
\toprule
{\bf Dataset}  & \multicolumn{1}{c}{Cora} & \multicolumn{1}{c}{Citeseer} & \multicolumn{1}{c} {Pubmed}  &\multicolumn{1}{c}{Actor}& \multicolumn{1}{c} {Chameleon}& \multicolumn{1}{c}{Squirrel}\\ \midrule 
{\bf{\#Nodes ($\boldsymbol{n}$)}}  & 2,708 & 3,327& 19,717  & 7,600 & 2,277 & 5,201\\
{\bf{\#Edges ($\boldsymbol{m}$)}} &  5,429 &  4,732& 44,338  & 26,659 & 31,371 & 198,353\\
{\bf \#Features ($\boldsymbol{d}$)} & 	1,433 & 3,703& 500 & 932 & 2,325 & 2,089 \\
{\bf \#Classes} & 7 & 6& 3& 5&  5& 5 \\
{\bf Homo. ratio ($\boldsymbol{h}$)}& 0.81 &  0.74 &  0.80 & 0.22 &  0.23 & 0.22  \\ \bottomrule
\end{tabular}
\end{table}

\spara{Datasets} Table~\ref{tbl:dataset} presents the detailed statistics of the six real-world datasets. The three homophily datasets, \ie {\em Cora}, {\em Citeseer}, and {\em Pubmed} are citation networks. Each graph node represents a research paper, and each edge denotes a citation relationship. Feature vectors of nodes are bag-of-words representations. The one-hot label assigned to each node stands for one research field of the paper. The rest three datasets are heterophily datasets. Specifically, {\em Actor} is a co-occurrence graph from the film-director-actor-writer network from WebKB3~\citep{TangSWY09,PeiWCLY20}. {\em Squirrel} and {\em Chameleon} are two datasets extracted from Wikipedia web pages, and nodes are categorized by the average amounts of monthly traffic~\citep{LiZCSLLQ22}. 

while the three heterophily datasets are Wikipedia datasets and WebKB3 dataset with homophily ratios around $0.22$. 

\spara{Running environment} All our experiments are conducted on a Linux machine with an RTX2080 GPU (10.76GB memory), Xeon CPU, and 377GB RAM.

\spara{Parameter settings} During the training process, learnable parameters are tuned with Adam~\citep{KingmaB14} optimizer. We set a patience of early stopping with $200$ epochs. For hyperparameters, we fix propagation hop $K=10$ for all tested datasets. The rest hyperparameters are selected in the following ranges. 

1. Learning rate: $[0.001, 0.005, 0.01, 0.05, 0.1, 0.15, 0.2]$;

2. Hidden dimension: $[64, 128, 256]$;

3. MLP layers: $[2,3,4]$;

4. Weight decays: $[0, 1e-4, 5e-4, 0.001]$;

5. Drop rates: $[0,0.1,0.2,\cdots, 0.9]$.

\spara{Choice of $\tau$} Ideally, the selection of $\tau$ for each dataset is highly related to its homophily ratio. Normally, we tune $\tau$ near the range of $h \pm 0.5$. In our experiments, the $\tau$ values are set as in Table~\ref{tbl:tauselection}.

\begin{table}[!t]
\centering
 \caption{$\tau$ Selection.} \label{tbl:tauselection}\vspace{-1mm} 
    \setlength{\tabcolsep}{0.5em}
 \small
\begin{tabular} {@{}c|ccccccc@{}}
\toprule
{\bf Dataset}  & \multicolumn{1}{c}{Cora} & \multicolumn{1}{c}{Citeseer} & \multicolumn{1}{c} {Pubmed}  &\multicolumn{1}{c}{Actor}& \multicolumn{1}{c} {Chameleon}& \multicolumn{1}{c}{Squirrel}\\ \midrule 
{\bf{$\tau$}}  & 1.0 & 0.9 & 0.8 & 0.1 & 0.7 & 0.7\\ \bottomrule
\end{tabular}
\end{table}

\revise{\subsection{Additional Experiments}\label{app:exp}

\spara{Homophily ratio estimation} For each dataset, we generate $10$ random splits of training/validation/test are generated. For each split, we estimate the homophily ratio from the training data, using it as the input to \ours to construct polynomial bases for node classification. Notice that we estimate a new homophily ratio for each split and then obtain the corresponding accuracy score. We present the estimated homophily ratio $\hat{h}$ with standard deviation averaged over the $10$ training sets for all datasets in Table~\ref{tbl:EstHomRatio}. In particular, we denote the averaged homophily ratio estimations as $\hat{h}_1$ and $\hat{h}_2$ for data splits settings of $60\%/20\%/20\%$ for polynomial filters and $48\%/32\%/20\%$ for model-optimized methods respectively. To verify the estimation difficulty, we vary the percentages of the training set in $\{10\%, 20\%, 30\%, 40\%, 50\%, 60\%\}$ on Cora and Squirrel, and then average the estimated homophily ratios $\hat{h}$ over $10$ random splits. The results are presented in Table~\ref{tbl:VaryEstHomRatio}.

\begin{table}[!t]
\centering
 \caption{Estimated homophily ratios $\hat{h}$.} \label{tbl:EstHomRatio}\vspace{-1mm} 
    \setlength{\tabcolsep}{0.5em}
 \small
 \renewcommand\arraystretch{1.3}
\begin{tabular} {@{}l|rrrrrr@{}}
\toprule
{\bf Dataset}  & \multicolumn{1}{c}{Cora} & \multicolumn{1}{c}{Citeseer} & \multicolumn{1}{c} {Pubmed} &\multicolumn{1}{c}{Actor}& \multicolumn{1}{c} {Chameleon}& \multicolumn{1}{c}{Squirrel}\\ \midrule 
{$\hat{h}_1$}  & 0.82 $\pm$ 0.01 & 0.70 $\pm$ 0.01 & 0.79 $\pm$ 0.005 & 0.21 $\pm$ 0.004 & 0.24 $\pm$ 0.01 & 0.22 $\pm$ 0.005 \\ \hline
{$\hat{h}_2$} & 0.82 $\pm$ 0.01 & 0.69 $\pm$ 0.014 & 0.79 $\pm$ 0.01 & 0.21 $\pm$ 0.004 &0.24 $\pm$ 0.01 & 0.22 $\pm$ 0.01 \\ \bottomrule
\end{tabular}
\end{table}


\begin{table}[!t]
\centering
 \caption{Estimated homophily ratio $\hat{h}$ over varying training percentages on Cora and Squirrel.} \label{tbl:VaryEstHomRatio}\vspace{-1mm} 
    \setlength{\tabcolsep}{0.5em}
 \small
 \renewcommand\arraystretch{1.3}
\begin{tabular} {@{}l|rrrrrr@{}}
\toprule
{\bf Dataset}  & \multicolumn{1}{c}{$10\%$} & \multicolumn{1}{c}{$20\%$} & \multicolumn{1}{c} {$30\%$} &\multicolumn{1}{c}{$40\%$}& \multicolumn{1}{c} {$50\%$}& \multicolumn{1}{c}{$60\%$}\\ \midrule 
{Cora}  & 0.83 $\pm$ 0.05 & 0.83 $\pm$0.04 & 0.83 $\pm$ 0.03 & 0.83 $\pm$ 0.01 & 0.82 $\pm$ 0.08 & 0.82$\pm$ 0.01 \\ \hline
{Squirrel} & 0.23 $\pm$  0.014 & 0.22 $\pm$  0.011 & 0.22 $\pm$  0.010 & 0.22 $\pm$  0.006 & 0.22 $\pm$  0.005 & 0.22 $\pm$ 0.005 \\ \bottomrule
\end{tabular}
\end{table}

As shown in Table~\ref{tbl:EstHomRatio}, the estimated values $\hat{h}_1$ and $\hat{h}_2$ derived from training datasets closely align with the actual homophily ratio $h$. There is a difference within a $2\%$ range across all datasets, except for Citeseer. As shown in Table~\ref{tbl:EstHomRatio}, the estimated homophily ratio $\hat{h}$ is approaching the true homophily ratio $h$ across varying percentages of training data. This observation verifies that a high-quality estimation of the homophily ratio is accessible by the training set.

\spara{Sensitivity of \ours to $\hat{h}$} We conduct an ablation study on the sensitivity of our proposed \ours to the estimated homophily ratio. Since \ours only utilizes the homophily basis on Cora ($\tau=1$ for Cora as shown in Table 5 in the Appendix in submission) and does not rely on the homophily ratio, we thus test \ours on Pubmed and Squirrel. To this end, we vary the estimated homophily ratio $\hat{h} \in \{0.78, 0.79, 0.80, 0.81, 0.82\}$ on Pubmed and  $\hat{h} \in\{0.19,0.20,0.21,0.22,0.23\}$ on Squirrel based on the results in Table~\ref{tbl:VaryEstHomRatio}. The achieved accuracy scores are presented in Table~\ref{tbl:SenseRatio}.

\begin{table}[!t]
\centering
 \caption{Estimated homophily ratio $\hat{h}$ over varying training percentages on Cora and Squirrel.} \label{tbl:SenseRatio}\vspace{-1mm} 
    \setlength{\tabcolsep}{0.5em}
 \small
 \renewcommand\arraystretch{1.3}
\begin{tabular} {@{}l|rrrrr@{}}
\toprule
{\bf Varying $\hat{h}$}  & \multicolumn{1}{c}{$0.78$} & \multicolumn{1}{c}{$0.79$} & \multicolumn{1}{c} {$0.80$} &\multicolumn{1}{c}{$0.81$}& \multicolumn{1}{c} {$0.82$}\\ \midrule 
{Squirrel}  & 91.03 $\pm$ 0.61 & 91.28 $\pm$ 0.64 & 91.34 $\pm$ 0.62 & 91.19 $\pm$ 0.67 & 91.17 $\pm$ 0.69 \\ \hline
{\bf Varying $\hat{h}$}  & \multicolumn{1}{c}{$0.19$} & \multicolumn{1}{c}{$0.20$} & \multicolumn{1}{c} {$0.21$} &\multicolumn{1}{c}{$0.22$}& \multicolumn{1}{c} {$0.23$}\\ \midrule 
{Pubmed}  & 66.22 $\pm$ 1.43 & 66.96 $\pm$ 1.38 & 67.09 $\pm$ 1.08 & 67.01 $\pm$ 1.25 & 66.69 $\pm$ 1.26 \\ \bottomrule
\end{tabular}
\end{table}

In Table~\ref{tbl:SenseRatio}, \ours demonstrates consistent accuracy scores across different homophily ratios for both Pubmed and Squirrel. Notably, the accuracy variations remain minor, staying within a $1\%$ range from the scores under the true homophily ratio, particularly on the Pubmed dataset.

\spara{Mitigation of Over-smoothing} As claimed, our proposed method \ours could address the over-smoothing problem. To verify in experiments, we generate a $1000$-length of homophily basis and our proposed heterophily basis on dataset Squirrel respectively, i.e., consisting of $1000$ basis vectors. We calculate the degrees of the angle formed by all two consecutive basis vectors and present the degree distribution in Table~\ref{tbl:oversmooth}. Notice that the degree is averaged across the dimension of node features.

\begin{table}[!t]
\centering
 \caption{Degree ($^\circ$) distribution of the homophily basis and \newbasis on Squirrel ($h=0.22$).} \label{tbl:oversmooth}\vspace{-1mm} 
 \resizebox{\columnwidth}{!}{%
    \setlength{\tabcolsep}{0.5em}
 \small
 \renewcommand\arraystretch{1.3}
\begin{tabular} {@{}l|rrrrrrrrr@{}}
\toprule
{\bf Basis}  & \multicolumn{1}{c}{$(v_1, v_2)$} & \multicolumn{1}{c}{$(v_2, v_3)$} & \multicolumn{1}{c} {$(v_3, v_4)$} &\multicolumn{1}{c}{$(v_4, v_5)$}& \multicolumn{1}{c} {$\cdots$} & \multicolumn{1}{c}{$(v_{996}, v_{997})$} & \multicolumn{1}{c}{$(v_{997}, v_{998})$} & \multicolumn{1}{c} {$(v_{998}, v_{999})$} &\multicolumn{1}{c}{$(v_{999}, v_{1000})$}\\ \midrule 
{Homo. Basis}  & $88.74^\circ$ & $87.99^\circ$ & $87.76^\circ$ & $86.51^\circ$ & $\cdots$ & $0.0123^\circ$ & $0.0118^\circ$ & $0.0115^\circ$ & $0.0114^\circ$ \\ \hline
{\newbasis}  & $69.72^\circ$ & $70.03^\circ$ & $70.01^\circ$ & $70.05^\circ$ & $\cdots$ & $71.37^\circ$ & $71.23^\circ$ & $71.32^\circ$ & $71.22^\circ$ \\ \bottomrule
\end{tabular}}
\end{table}

As shown, degrees of the formed angles by two consecutive basis vectors from the homophily basis approach to $0^\circ$, which indicates that the homophily basis on Squirrel converges asymptotically. On the contrary, degrees of our basis keep around $71^\circ$  determined by our setting $\theta=\tfrac{\pi}{2}(1-h)$ where $h=0.22$ for Squirrel. Notice that the variation of degrees is due to the average across $2089$-dimension features of Squirrel. 

}

\end{sloppy}
\end{document}